\documentclass[preprint,12pt]{elsarticle}

\usepackage{amsmath,amssymb,amsfonts}
\usepackage{algorithmic}
\usepackage{graphicx}
\usepackage{textcomp}
\usepackage{comment}
\usepackage{multirow} 
\usepackage{float}
\usepackage{booktabs}
\usepackage{subcaption} 
\usepackage{hyperref}
\usepackage{placeins} 
\usepackage{siunitx} 
\usepackage{amsthm}
\usepackage{bm}
\usepackage{a4wide}

\theoremstyle{definition}
\newtheorem{definition}{Definition}

\begin{document}

\begin{frontmatter}

\title{Data-Driven Temperature Modelling of Machine Tools by Neural Networks: A Benchmark}

\author[hsu]{C. Coelho\corref{cor1}}
\ead{cecilia.coelho@hsu-hh.de}

\author[hsu]{M. Hohmann}
\author[tud]{D. Fernández}
\author[tud]{L. Penter}
\author[tud]{S. Ihlenfeldt}
\author[hsu]{O. Niggemann}

\address[hsu]{Institute for Artificial Intelligence, Helmut Schmidt University, 22043 Hamburg, Germany}
\address[tud]{Chair of Machine Tools Development and Adaptive Controls, Dresden University of Technology TUD, 01069 Dresden, Germany}

\cortext[cor1]{Corresponding author}

\begin{abstract}
Thermal errors in machine tools significantly impact machining precision and productivity. Traditional thermal error correction/compensation methods rely on measured temperature-deformation fields or on transfer functions.
Most existing data-driven compensation strategies employ neural networks (NNs) to directly predict thermal errors or specific compensation values. While effective, these approaches are tightly bound to particular error types, spatial locations, or machine configurations, limiting their generality and adaptability. In this work, we introduce a novel paradigm in which NNs are trained to predict high-fidelity temperature and heat flux fields within the machine tool. The proposed framework enables subsequent computation and correction of a wide range of error types using modular, swappable downstream components. The NN is trained using data obtained with the finite element method under varying initial conditions and incorporates a correlation-based selection strategy that identifies the most informative measurement points, minimising hardware requirements during inference. We further benchmark state-of-the-art time-series NN architectures, namely Recurrent NN, Gated Recurrent Unit, Long-Short Term Memory (LSTM), Bidirectional LSTM, Transformer, and Temporal Convolutional Network, by training both specialised models, tailored for specific initial conditions, and general models, capable of extrapolating to unseen scenarios. The results show accurate and low-cost prediction of temperature and heat flux fields, 
laying the basis for enabling flexible and generalisable thermal error correction in machine tool environments. 
\end{abstract}


\end{frontmatter}

\newpage

\section{Introduction} \label{sec:introduction}

With the rise of the need for high-precision manufacturing, aimed at reducing costs and wasted resources, the focus on thermal error mitigation in machine tools has intensified. Thermally induced deformations represent one of the most significant sources of machining inaccuracies, accounting for up to 75\% of the total positioning errors in computer numerical control machine tools, making their prediction and compensation critical for ensuring manufacturing precision \cite{gebhardt2014high,mayr2012thermal}. The remaining sources of error, however, are considerably more difficult to capture, as they often arise from non-linear boundary conditions (such as coolant behaviour, air flow variations, and other environmental interactions) that introduce complex and highly coupled dynamics, challenging the development of accurate and comprehensive error models.

\begin{figure*}[!t]
        \centering
        \includegraphics[width=\textwidth]{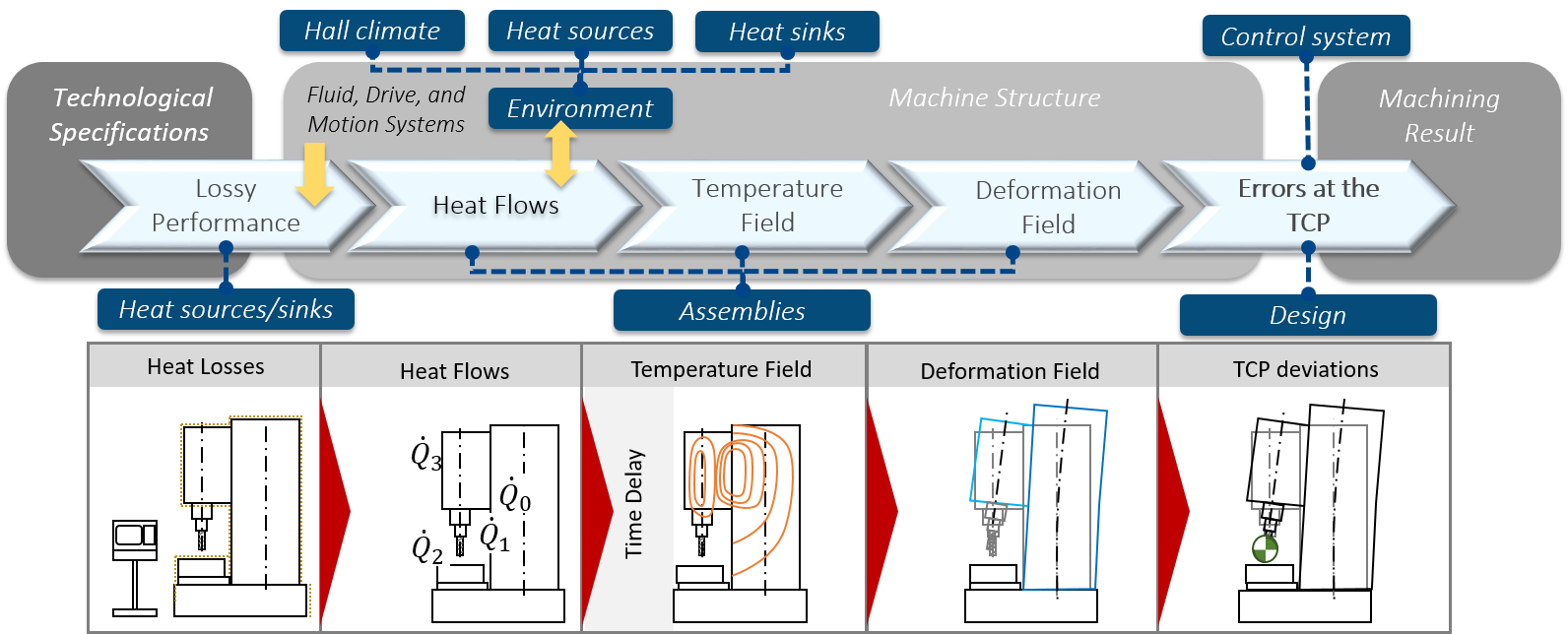}
        \caption{Process chain: from heat losses to errors at the TCP. Adapted from \cite{ihlenfeldt2020adjustment}.}
        \label{fig:mainscheme_introduction}
    \end{figure*}

While hardware-based passive thermal design strategies (such as adopting cooling systems, thermal isolation and materials with low thermal expansion coefficients) can reduce thermal induced errors, they are often expensive and may not be sufficient to maintain the needed manufacturing precision requirements. In the other hand, software-based strategies are capable of real-time thermal error compensation, being able to adapt to varying operating conditions in a cost-effective way \cite{chen1995computer,khan2011methodology}.
Predicting the thermal behaviour of machine tools is a complex task since the different structures are subject to various heat sources during operation, including friction from bearings and guideways, motor heat dissipation, heat released during the cutting process, and environmental temperature variations. These different heat sources create non-uniform temperature distributions that evolve over time, causing structural deformations that directly affect the relative positioning between the cutting tool and workpiece \cite{9047921,wu2003thermal,li2019thermal}, Figure \ref{fig:mainscheme_introduction}.

Traditional methods 
 on modelling thermally induced errors in machine tools
 rely primarily on the Finite Element Method (FEM) simulations using Modelling Order Reduction (MOR) techniques 
  \cite{tanabe2024development,li2019reliability} and empirical regression models \cite{cao2022method,lian2022novel}. Approaches based on FEM are able to provide detailed and accurate behaviours by discretising machine tool structures and solving heat transfer governing equations. However, to achieve high precision, these suffer from high, and often prohibitive, computational costs that make real-time implementation impractical \cite{marinkovic2019survey}. The usage of MOR coupled with FEM allows to achieve real-time predictive models but expert knowledge is needed in the process.
Contrarily, linear regression \cite{cao2022method,lian2022novel} and transfer function models \cite{polyakov2019simulation,zhang2017thermal} offer computational efficiency but have limited ability to capture complex non-linear and time-dependent thermal behaviours inherent in machine tools. Furthermore, these traditional methods face challenges in generalising across different operating conditions and require extensive recalibration for varying thermal scenarios \cite{inigo2021digital}.

Neural networks (NNs) have shown promising results in addressing challenges in multiple fields, previously dominated by traditional methods, driving significant research interest for thermal error correction in machine tools. 
For instance, works such as \cite{gao2023modeling,bao2023thermal,de2022development,liu2021thermal,mu2025convolutional,li2024thermal} propose training NNs on historical thermal error measurements to model their evolution over time, enabling prediction of future errors and providing a foundation for real-time correction strategies. 
In contrast, \cite{liu2021thermalBayes} focus on training a NN to directly output the required error correction from thermal error measurements, thus bypassing the conventional two step process of error prediction followed by correction computation. Also, as to predict absolute temperature, \cite{boos2022improving} uses NNs trained with a mix of real data and synthetic data from domain-randomised MOR FE simulation models.


While much of the research thus far has emphasised predicting and correcting thermal errors, directly, using NNs, a less explored but promising direction is data-driven modelling of thermal fields within machine tools. NN-based temperature modelling seeks to learn the complex, non-linear and dynamic evolution of temperature distributions from a dataset, thus reducing the need for real-time measurements and being able to generalise across varying operational and environmental conditions. 
To the best of our knowledge, no prior work has addressed the data-driven modelling of spatial and temporal temperature distributions within machine tools using NN architectures.


In this work, we present the first study on NN-based thermal field and heat flux modelling for machine tools, shifting from error compensation to thermal state prediction. This paradigm change makes our approach inherently more generalisable across machine configurations, sensor layouts, and operational regimes since thermal fields and heat fluxes arise from fundamental thermodynamic processes that are governed by universal physical laws and are less dependent on the specific geometric characteristics of a given machine tool. Thermal errors can be obtained as well-defined, often linear, functions of the thermal field and known structural properties, thus a learnt temperature model can be reused to compute different types of errors, such as volumetric, or to accommodate new sensor configurations with minimal adaption. Decoupling thermal state estimation and error computation enables high-resolution temperature data from a NN to serve as input for computation and compensation algorithms or real-time optimal control approaches, allowing efficient, robust, and transparent error correction strategies without the computational burden of FEM simulations. Leveraging this, we additionally propose a novel modular framework for thermal error correction in machine tools that uses the predicted thermal fields as input to a swappable error computation module that is connected to a swappable error correction module. 
To the best of our knowledge, this is the first time such an approach for thermal error correction in machine tools is proposed.


Furthermore, a major limitation of direct FEM-based thermal field and heat flux modelling emerges from the computational cost of simulating high-resolution meshes and the 
fact that non-linearities, variability of environmental effects and the optimisation of thermal performance (such as novel cooling strategies using computational
fluid dynamics) make the models more complex, even when using MOR.
To address this, we propose a two-stage sensor reduction and reconstruction approach by analysing the correlation between nodes and only using the most informative. The excluded nodes are then reconstructed with linear fitting. 

Additionally, we present a comprehensive benchmark for data-driven temperature modelling of machine tools using various state-of-the-art NN architectures. Five different NN, time-series capable, architectures are used and their performance compared, namely Recurrent Neural Network (RNN) Gated Recurrent Units (GRU), Long-Short Term Memory (LSTM), Bidirectional LSTM (BiLSTM), Transformer, and Temporal Convolutional Networks (TCN), at modelling a FEM-generated dataset.

In summary, the main contributions of this work are:
\begin{enumerate}
    \item \textbf{Data-driven Temperature Modelling for Machine Tools:} this work is the first to directly model the thermal field and heat flux over time using a NN, moving beyond conventional error-centric approaches to establish a new approach to thermal error correction. Furthermore, we present a novel modular thermal error correction framework with three main components: a temperature predictor (thermal field and heat flux), an error computation module, and an error correction module;

    \item \textbf{Improved Generalisability:} thermal fields and heat fluxes can be used to compute various types of errors for multiple points or subsystems. Conversely, training models directly on error prediction bound them to a particular error type, location, or operating condition;

     \item \textbf{Sensor-efficient Node Selection:} this work is the first to address the impracticality of using a large number of sensors implied by high-resolution FEM discretisation, using NN. We propose a two-stage node selection strategy that uses a correlation matrix to identify and disregard nodes with low information contribution. Then the NN is trained with the nodes left while the omitted nodes are efficiently reconstructed based on similarity using a proposed linear regression approach;
    

    \item \textbf{Systematic Architecture Comparison:} a thorough comparison of RNN, GRU, LSTM, BiLSTM, Transformer, and TCN models for temperature modelling is presented, providing guidance on architecture selection in these applications.
\end{enumerate}

The rest of the paper is organised as follows. Section \ref{sec:back} provides a brief overview on thermal errors in machine tools and the NN architectures used in this work. In Section \ref{sec:method}, the approach for temperature modelling with NNs is discussed. Section \ref{sec:exps} details the benchmarking setup and discusses the results. To finish, the conclusions are presented in Section \ref{sec:conc}. Additionally, 

\section{Background} \label{sec:back}

\subsection{Thermal Errors in Machine Tools}

Thermal errors in machine tools originate from technological specifications and environmental influences (ambient air temperature, solar radiation, radiators, adjacent machinery, foundation heat, open hall conditions, and door effects), combined with internal heat sources and sinks (e.g., motors, spindles) \cite{Mayr2012}. These effects induce performance losses, particularly pronounced in fluid, drive, and motion systems, which generate heat and disrupt thermal equilibrium.

The precision of machine tools is deeply impacted by thermal effects, which induce displacements along machine axes 
and a deformation field in the machine tool
due to the heating of components and its assemblies, from both internal sources and external environmental factors. 
The primary objective of integrating thermal compensation into machine tools is to enhance production accuracy in an energy-efficient and cost-effective manner \cite{grossmann2015thermo}. This can be achieved either by incorporating predicted thermal drift as corrections within the Numeric Control (NC) program (offline compensation) or by applying real-time position offsets during machining (online compensation) \cite{tseng1997real}.
These approaches can be categorized into compensation and correction methods: compensation focuses on reducing or preventing thermally induced deformations by managing heat inputs and flow, whereas correction addresses displacements at the Tool Centre Point (TCP) through adaptive movement specifications. Errors at TCP may emerge due to design or control system limitations, manifesting as motion errors and structural distortions that contribute to TCP deviations.


The governing equation for lossy performance, in this context, is given by: 
    
    \begin{equation}
    \dot{Q}_{\text{loss}} = \sum_{b=1}^{N_{\text{gen}}} \dot{Q}^{\text{gen}}_b - \sum_{v=1}^{N_{\text{diss}}} \dot{Q}^{\text{diss}}_v,
    \end{equation}
    
    \noindent where $\dot{Q}^{\text{gen}}_b$ are the contributions from heat sources ($b=1 \dots, N_{\text{gen}}$), such as motor losses and friction, and $\dot{Q}^{\text{diss}}_v$ are contributions from heat dissipation mechanisms ($v=1 \dots, N_{\text{diss}}$), through convection and radiation.

    Heat conduction within the machine tool follows:
    
    
    \begin{equation}
        \varrho c_p \dfrac{\partial T}{\partial t} = \nabla  (k \nabla T) + Q_{\text{int}},
    \end{equation}

\noindent where \( \varrho \) is the density, \( c_p \) is the specific heat capacity, \( T \) the thermal field, \( k \) the thermal conductivity, and \( Q_{\text{int}} \) the internal heat generation (e.g., from spindle operation). 

Convective and radiative heat transfer are modelled as:

    \begin{equation}
        q_{\text{conv}} = \emph{h} A (T_s - T_\infty),
    \qquad
        q_{\text{rad}} = \varepsilon \varsigma A (T_s^4 - T_{\text{surr}}^4),
    \end{equation}
    
    \noindent where \( \emph{h} \) is the convective heat transfer coefficient, \( A \) the surface area, \( \varepsilon \in [0, 1] \) the emissivity, and \( \varsigma \) the Stefan-Boltzmann constant ($5.67 \times 10^{-8}  $W/(m$^2$K$^4$). Here, $T_s$ is the surface temperature, $T_\infty$ the ambient air temperature, and $T_{\text{surr}}$ the surrounding temperature.
    
    The thermal field induces expansion governed by:
    
    \begin{equation}
        \varepsilon_{\text{th}} = \alpha \Delta T \mathbf{I},
        \label{4}
    \end{equation}

   \noindent where \( \alpha \) is the coefficient of thermal expansion,  $\mathbf{I}$ as the identity tensor and \( \Delta T \) the temperature change, leading to deformations in machine components. 
   
   The deformation field satisfies the structural equilibrium equation:
   
    \begin{equation}
        \nabla  \boldsymbol{\pi}_{th} + \boldsymbol{F} = \boldsymbol{0},
        \label{5}
    \end{equation}

    \noindent with thermal stress contribution,

    \begin{equation}
        \boldsymbol{\pi}_{\text{th}} = E \alpha \Delta T,
        \label{6}
    \end{equation}
    
   \noindent where \( E \) is the Young’s modulus), $\boldsymbol{\pi} \in \mathbb{R}^{3 \times 3}$ the stress tensor, and $\boldsymbol{F} \in \mathbb{R}^3$ the external force field.

\subsubsection{Types of Thermally Induced Errors}

Thermally induced errors in machine tools can be systematically described using the thermo-elastic functional chain illustrated in Fig. \ref{fig:mainscheme_introduction}. Heat is generated within the machine due to internal sources (e.g., spindle motors, bearings, frictional losses) and is dissipated through sinks (e.g., coolant systems, ambient air exchange). The resulting heat flows propagate through machine components and assemblies, leading to a non-uniform temperature distribution across the structure.
This temperature field induces thermal expansion and contraction of the machine elements, thereby generating a deformation field. The combined effect of these thermo-elastic deformations manifests as deviations in the TCP. In essence, thermo-elastic deformation is the primary physical mechanism responsible for thermally induced errors at the TCP.
These errors can be categorized into two main types: positional errors (thermal drift) and orientation errors (angular deviations). When considered together, they constitute the volumetric error, which is defined as the spatial displacement of the TCP relative to its intended position.

\paragraph{Thermo-elastic Deformations}

Thermo-elastic deformations arise from the fundamental coupling between thermal fields and mechanical stress-strain states within machine structures, which leads to complex displacement patterns \cite{naumann2016computation}.

The thermo-elastic equilibrium equation can be presented as follows,


    \begin{equation}
        \nabla \left[ \mathbf{C} : \left( \boldsymbol{\varepsilon} - \boldsymbol{\alpha} \Delta T \mathbf{I} \right) \right] + \mathbf{F} = 0.
        \label{equilibrium}
    \end{equation}


Equation \eqref{equilibrium} governs thermo-elastic deformations in machine tools due to temperature variations. Here, $\mathbf{C}$ is the elasticity tensor describing material stiffness, $\boldsymbol{\varepsilon} \in \mathbb{R}^{3\times 3}$ is the total strain tensor measuring overall deformation, $\alpha \Delta T \mathbf{I}$ is the thermal strain from \eqref{4}, $\boldsymbol{\varepsilon}_{\text{th}} = \alpha \Delta T$, with $\alpha$ as the thermal expansion coefficient, $\Delta T$ as the temperature change, and $\mathbf{I}$ as the identity tensor, while $\mathbf{F}$ represents body forces like gravity. Derived from continuum mechanics, it extends eq.\eqref{5} by using the stress $ \boldsymbol{\sigma} = \mathbf{C} : \left( \boldsymbol{\varepsilon} - \alpha \Delta T \mathbf{I} \right) $. This accounts for isotropic expansion and complex stress states, ensuring tensor compatibility for 3D deformations.

It is then possible to quantify thermal-induced positional and orientation errors by solving the thermo-elastic equilibrium equation and mapping deformation to the TCP \cite{Brecher2021}. These models form the basis for error compensation strategies, which may include pre-emptive correction via Computerised Numerical Control (CNC) offsets \cite{Brecher2021}, closed-loop control using temperature sensors \cite{inigo2024digital}, or integration into digital twins for adaptive, real-time compensation \cite{colinas2023thermal, inigo2024digital}.

\paragraph{Position Drift}

Position drift, also known as thermal drift refers to the slow, time-dependent change in the nominal position of a machine tool's structural elements (such as spindle, tool, or workpiece) caused primarily by thermally induced deformations \cite{Mayr2012}. This effect results form the accumulation of heat generated by internal sources (spindle, motors, bearings, drives, friction) and external environmental factors (ambient air temperature fluctuations, solar radiation, airflow, etc.) and manifests as an undesired displacement of the TCP relative to its reference configuration. Unlike volumetric errors, which encompass all geometric and kinematic deviations, position drift specifically refers to the offset error measured at the TCP. Although it is not governed by a single equation, its quantitative description derives from the general thermo-elastic equilibrium, which governs the relationship between temperature, stress, and deformation. 
Position drift in CNC machine tools, primarily caused by thermal and mechanical effects, has been addressed through various modelling and compensation strategies.

Thus, the displacement field $\mathbf{u}(\mathbf{x},t) \in \mathbb{R}^3$ of the machine structure can be obtained from the governing equation \eqref{equilibrium}. Since strain is related to displacement by
\begin{equation}
\boldsymbol{\varepsilon} = \tfrac{1}{2} \big( \nabla \mathbf{u} + (\nabla \mathbf{u})^\top \big),
\end{equation}
solving the thermo-elastic problem provides $\mathbf{u}(\mathbf{x},t)$ at every point $\mathbf{x}$ and time $t$. The thermal drift relevant for machining accuracy is then expressed as the displacement at the TCP,
\begin{equation}
\Delta \mathbf{r}_{TCP}(t) = \mathbf{u}(\mathbf{x}_{TCP},t),
\end{equation}
which directly quantifies the undesired positional error caused by thermal deformations. This formulation establishes the link between the fundamental thermo-elastic model and the practical definition of thermal drift in machine tools.

 For large gantry machines, the Environmental Temperature Consideration Prediction model combines time series analysis, Fourier synthesis, and regression to predict thermal errors from both environmental and internal heat sources, enabling pre-emptive error compensation \cite{tan2014thermal}. In feed axis systems under thermal–mechanical loads, radial basis function neural networks have been employed to track, decompose, and predict position errors, allowing accurate assessment of machine status and source tracing of drift \cite{wang2025position}. For motorized spindles, time series modelling of thermal expansion using sensor measurements and Yule-Walker equations supports real-time error compensation, reducing axial deviations significantly and improving machining accuracy \cite{yang2014thermal}. Collectively, these approaches demonstrate that predictive modelling combined with compensation strategies is effective for mitigating position drift in CNC machining.

\paragraph{Orientation and Volumetric Errors}

Orientation errors manifest as angular deviations of the TCP due to non-uniform thermal expansion and differential heating of machine structural elements from asymmetric thermal fields \cite{yang2015thermal}. 

Volumetric errors represent the combined spatial distribution of all thermal distortions throughout the machine, exhibiting strong position-dependence. These errors have a motion of six degrees-of-freedom (three translations and three rotations) at any point within the machine structure \cite{krulewich1998rapid,brecher2015volumetric,brecher2016volumetric}.

Governing equations for thermally induced volumetric errors may differ based on the approach. In rigid-body kinematics-based methods, volumetric errors are expressed by linking axis-wise geometric deviations with TCP displacements. For example, \cite{chen2022modeling} represents the positional deviations $\Delta${x}, $\Delta${y} and $\Delta${z} in terms of elementary axis errors. While \cite{aguado2019study} extends this formulation by incorporating thermal expansion directly into the kinematic chain. Including the uncertainty of the Coefficient of Thermal Expansion (CTE), one axis-wise expression is given by:

    \begin{equation}
        \overline{X} = 
               -\Big[ \text{x} \big( 1 + (\alpha_{\text{x}} + u(\alpha_{\text{x}})) \cdot (T_i - T_{20}) \big) \Big] + \delta_{\text{x}}(\text{x}),
    \end{equation}

\noindent where $\alpha_{\text{x}}$ is the CTE of the x-direction structural element, $u(\alpha_{\text{x}})$ its uncertainty, $T_i$ -$T_{20}$ the temperature deviation from a reference temperature of 20$^{\circ}$C, and $\delta_{\text{x}}({\text{x}})$ the geometric positioning error associated with the x-axis. Analogous formulations apply for $\overline{Y}$ and $\overline{Z}$ \cite{aguado2019study}.

In practise, volumetric error prediction often uses empirical or data-driven models, such as multiple linear regression models, or autoregressive models with exogenous inputs models \cite{inigo2024digital}, or transfer functions mapping temperatures, motor loads, or spindle speeds to TCP displacements \cite{ngoc2023deep}.

State-of-the-art compensation strategies typically combine physics-based rigid-body kinematic models augmented with thermal expansion terms, and data-driven models fitted to sensor measurements for real-life adaption. Physics-based thermal models enhance interpretability and ensure thermo-mechanical consistency but are most effective when integrated into hybrid frameworks linking distributed heat-induced distortions with global volumetric error \cite{xiang2015thermal,wu2025knowledge}.

\subsection{Neural Networks for Time-series}

Time-series are ordered, time-indexed sequences of data $\boldsymbol{X} = \{\boldsymbol{x}_1, \dots, \boldsymbol{x}_N\}$ (with $\boldsymbol{x}_i \in \mathbb{R}^d$ and $i \in 1, \dots, N$), that allow modelling temporal dynamics in systems such as machine tools. The correspondent target values are denoted $\boldsymbol{Y} = \{\boldsymbol{y}_1, \dots, \boldsymbol{y}_N\}$, where $\boldsymbol{y}_i$ represents the expected output at time step $i$. To capture their temporal dependencies and complex relationships, special designed NN architectures have to be employed \cite{bishop2023deep}. Unlike \emph{classical} NNs, these process sequential data step-by-step or in parallel, retaining past steps in memory to predict future ones. RNNs \cite{rumelhart1985learning} and their more advanced variants, such as GRU \cite{chung2014empirical} and LSTM \cite{hochreiter1997long}, incorporate gating mechanisms that enable learning of short- and long-term dependencies. Furthermore, bidirectional variants \cite{schuster1997bidirectional}, such as BiLSTM \cite{graves2005framewise}, consider information flow in both directions, forward and backward in time, allowing the extraction of more complex patterns from data. Transformers \cite{vaswani2017attention} leverage a self-attention mechanism to improve performance in capturing long-term dependencies and allow training to be parallelised.

\subsubsection{Recurrent Neural Networks}

RNNs were introduced as the pioneer NN architecture capable of handling sequential data by incorporating feedback loops that allow information from previous time steps to influence current computations. Unlike \emph{classical} NNs that process inputs independently, RNNs keep a hidden state $\boldsymbol{h}_i \in \mathbb{R}^n$, at each time step $i$, that retains information from all previous time steps allowing the network to model temporal dependencies and sequential patterns in the data, simulating memory:

\begin{equation}
   h_{i} = \sigma(\boldsymbol{W}_{\text{feedback}} \boldsymbol{h}_{i-1} +  \boldsymbol{W}_{\text{input}}\boldsymbol{x}_{i} + \boldsymbol{b}_i),
\end{equation}

\noindent where $\sigma$ is an activation function,  $\boldsymbol{W}_{\text{input}} \in \mathbb{R}^{n \times d}$ transforms the current input, 
$\boldsymbol{W}_{\text{feedback}} \in \mathbb{R}^{n \times n}$ processes the previous hidden state $\boldsymbol{h}_{i-1}$, and $\boldsymbol{b}_i \in \mathbb{R}^n$ is the bias vector. The initial hidden state $\boldsymbol{h}_0$ needs to be initialised. The output prediction for a time step $i$ is given by $\boldsymbol{\hat{y}}_i = \sigma(\boldsymbol{W}_{\text{output}}\boldsymbol{h}_{i} + \boldsymbol{b}_{\text{output}})$, where $\boldsymbol{W}_{\text{output}}$ transforms the hidden state into the output space \cite{rumelhart1985learning}. 

RNNs share the parameters $\boldsymbol{W}_{\text{input}}, \boldsymbol{W}_{\text{feedback}}$ and $\boldsymbol{W}_{\text{output}}$ across all time steps, enabling them to handle sequences of variable length efficiently \cite{rumelhart1985learning}. However, due to this reason, RNNs suffer from the vanishing and exploding gradient problem, restricting their learning ability \cite{coelho2024enhancing,pascanu2013difficulty}.

\subsubsection{Long-Short Term Memory}

LSTMs were presented to address the limitations of RNNs by incorporating a gating mechanism that controls information flow. LSTMs leverage feedback loops by using an additional internal memory cell state $C_i \in \mathbb{R}^n$, that acts as a long-term memory, and three gates: the input gate $I_i  \in \mathbb{R}^n$ controls which new information to keep; the forget gate vector $F_i  \in \mathbb{R}^n$ controls what information to discard; the output gate vector $O_i \in \mathbb{R}^n$) controls the output. The computations are as follows:

\begin{equation}\label{eq:gates}
\begin{array}{rcl}
   I_{i}&=  &   \sigma( \boldsymbol{W}_{xin}\boldsymbol{x}_i +  \boldsymbol{W}_{hin}\boldsymbol{h}_{i-1}+\boldsymbol{b}_{in})\\
   F_{i}&= &\sigma(  \boldsymbol{W}_{xf}\boldsymbol{x}_i+ \boldsymbol{W}_{hf}\boldsymbol{h}_{i-1} +\boldsymbol{b}_f) \\
    O_i&=&\sigma( \boldsymbol{W}_{xo}\boldsymbol{x}_i +  \boldsymbol{W}_{ho}\boldsymbol{h}_{i-1}+\boldsymbol{b}_o) \\
    \tilde{C}_{i} &=& \tanh(\boldsymbol{W}_{xc} \boldsymbol{x}_i  + \boldsymbol{W}_{hc} \boldsymbol{h}_{i-1} + \boldsymbol{b}_c) \\
    C_{i}&=& F_i \odot C_{i-1}+I_i \odot \tilde{C}_i  \\
     \boldsymbol{h}_{i}&=& O_{i} \odot \tanh(C_i),
\end{array}
\end{equation}

\noindent where $\sigma$ denotes the sigmoid function, $\odot$ denotes the Hadamard product, $\tilde{C}_i \in \mathbb{R}^n$ is the candidate memory, $\boldsymbol{W}_{xin},\boldsymbol{W}_{xf},\boldsymbol{W}_{xo} \in \mathbb{R}^{n \times d}$, $\boldsymbol{W}_{hin},\boldsymbol{W}_{hf},\boldsymbol{W}_{ho} \in \mathbb{R}^{n \times n}$, and $\boldsymbol{b}_{in}, \boldsymbol{b}_{f}, \boldsymbol{b}_{o} \in \mathbb{R}^n$. The initial memory cell $C_0$ needs to be initialised \cite{hochreiter1997long}.

\subsubsection{Gated-Recurrent Unit}

Motivated by the reducing the computational cost of LSTMs, GRUs simplify the LSTM architecture by using two gates instead of three: the reset gate $R_i \in \mathbb{R}^n$ controls what information to discard; the update gate $Z_i \in \mathbb{R}^n$ controls what past information to keep and what new information to add. The computations are as follows:

\begin{equation}\label{eq:gates}
\begin{array}{rcl}
   R_{i}&= &\sigma(\boldsymbol{W}_{xr}\boldsymbol{x}_i+ \boldsymbol{W}_{hr}\boldsymbol{h}_{i-1} +\boldsymbol{b}_r) \\
   Z_{i}&= &\sigma(\boldsymbol{W}_{xz}\boldsymbol{x}_i+ \boldsymbol{W}_{hz}\boldsymbol{h}_{i-1} +\boldsymbol{b}_z) \\
   \boldsymbol{\tilde{h}}_{i} &=& \tanh(\boldsymbol{W}_{xh} \boldsymbol{x}_i  + \boldsymbol{W}_{h\tilde{h}}(R_i \odot \boldsymbol{h}_{i-1}) + \boldsymbol{b}_{\tilde{h}}) \\
   \boldsymbol{h}_i &=& (1-Z_i) \odot \boldsymbol{h}_{i-1} + Z_i \odot \boldsymbol{\tilde{h}}_i,
\end{array}
\end{equation}

\noindent where $\boldsymbol{\tilde{h}}_i \in \mathbb{R}^n$ is the candidate hidden state, $\sigma$ is the sigmoid function,  $\boldsymbol{W}_{xr},\boldsymbol{W}_{xz} \in \mathbb{R}^{n \times d}$, $\boldsymbol{W}_{hr},\boldsymbol{W}_{hz}, \boldsymbol{W}_{h\tilde{h}} \in \mathbb{R}^{n \times n}$, and $\boldsymbol{b}_{r}, \boldsymbol{b}_{z}, \boldsymbol{b}_{\tilde{h}} \in \mathbb{R}^n$ \cite{chung2014empirical}.

\subsubsection{Bidirectional Long-Short Term Memory}

BiLSTM extends LSTM by processing sequences in both forward and backward directions, being able to extract dependencies forwards and backwards in time. A BiLSTM is composed of two separate LSTM blocks stacked in parallel: a forward LSTM, $\overrightarrow{\text{LSTM}}$, that processes the sequences from $i=1$ to $i=N$; a backward LSTM, $\overleftarrow{\text{LSTM}}$, that processes the sequences from $i=N$ to $i=1$. The final hidden state $\boldsymbol{h}_i \in \mathbb{R}^{2n}$, used to compute the network prediction, is formed by concatenating the forward $\boldsymbol{\overrightarrow{h}}_i  \in \mathbb{R}^n$ and the backward $\boldsymbol{\overleftarrow{h}}_i  \in \mathbb{R}^n$ hidden states \cite{graves2005framewise}.

\subsubsection{Transformer}

Transformers are a revolutionising architecture that replace the recurrence mechanism with self-attention, enabling parallel processing and capturing long-range dependencies better than the previous seen approaches. This is done by processing all sequence elements simultaneously using a multi-head self-attention mechanism that uses a scaled dot-product to compute attention weights between all pairs of positions in the sequence. Considering an input sequence represented as queries $\boldsymbol{Q}$, keys $\boldsymbol{K}$, and values $\boldsymbol{V}$, obtained through learnt linear transformations:

\begin{equation}
\begin{array}{rcl}
    \boldsymbol{Q} &=& \boldsymbol{XW}^Q \\
    \boldsymbol{K} &=& \boldsymbol{XW}^K \\
    \boldsymbol{V} &=& \boldsymbol{XW}^V \\
    \text{Attention}(\boldsymbol{Q}, \boldsymbol{K}, \boldsymbol{V}) &=& \text{softmax}\left( \dfrac{\boldsymbol{QK^T}}{\sqrt{d_k}} \right) \boldsymbol{V},
\end{array}
\end{equation}

\noindent with $d_k$ the dimension of the key vectors, $\sqrt{d_k}$ is a scaling factor to prevent the softmax function from saturating, and $\boldsymbol{W}^Q, \boldsymbol{W}^K, \boldsymbol{W}^V \in \mathbb{R}^{d \times d_k}$ are trainable parameter matrices \cite{vaswani2017attention}.

Multi-head attention extends this idea by running $H$ attention heads in parallel, each with different learnt projections, thus capturing various types of relationships simultaneously:

\begin{equation}
    \text{Multi-head}(\boldsymbol{Q}, \boldsymbol{K}, \boldsymbol{V}) = \text{concat}(\text{head}_1, \dots, \text{head}_H) \boldsymbol{W}^O,
\end{equation}

\noindent where $\text{head}_i = \text{Attention}(\boldsymbol{XW}^Q_i, \boldsymbol{XW}^K_i, \boldsymbol{XW}^V_i)$, and $\boldsymbol{W}^O \in \mathbb{R}^{(H \times d_{\text{head}}) \times {d_{\text{embedding}}}}$, with $d_{\text{head}}$ the dimension of the output of each attention head and ${d_{\text{embedding}}}$ the dimension of the representation throughout the transformer \cite{bishop2023deep}. The multi-head attention is then combined with position-wise feed-forward networks to refine representations, and residual connections to stabilise training and preserve information flow \cite{vaswani2017attention}.

\subsubsection{Temporal Convolutional Networks}

TCNs were introduced to provide a convolution-based approach to sequence modelling that captures long-range dependencies without using recurrence. The convolutional layers allow the processing of entire sequences in parallel while maintaining temporal causality \cite{bai2018empirical,lea2017temporal}.

TCNs rely on two key concepts, causal convolutions and dilated convolutions. Causal convolutions insure that the output at a time step $i$ depends only on inputs from previous time steps, ensuring temporal causality. 
Considering a 1-dimensional (1D) causal convolution with kernel size $k+1$, the output at time $i$ is given by:

\begin{equation}
\label{eq:causalCon}
    \boldsymbol{s}_i = \sum_{j=0}^k \boldsymbol{W}_j \boldsymbol{x}_{i-j},
\end{equation}

with $\boldsymbol{W}_j$ the trainable convolutional weights and $\boldsymbol{x}_{i-j}$ the past inputs. A limitation of causal convolutions is that the range of input steps that influence an output is determined by the kernel size and the network depth. Thus, for a kernel for size $k+1$ and a stack of $L$ layers the covered range is $1+kL$. For long sequences this number grows linearly with depth so, to model dependencies across a big number of time steps, an impractically large number of layers would be needed. To address this problem, TCNs use dilated convolutions \cite{bai2018empirical}.

In a dilated convolution the kernel skips input positions according to a dilation factor $d$. Thus, rewriting \eqref{eq:causalCon} gives:

\begin{equation}
    \boldsymbol{s}_i = \sum_{j=0}^k \boldsymbol{W}_j \boldsymbol{x}_{i-dj}.
\end{equation}

\noindent To stabilise training and allow deeper networks, TCNs stack multiple layers of dilated causal convolutions inside residual blocks \cite{bai2018empirical}. 
Since gradients propagate through network depth rather than time steps, TCNs prevent the vanishing and exploding gradient problem, common in RNNs. Also, TCNs may provide longer effective memory and more stable training than LSTM and GRU \cite{bai2018empirical}. Comparing with transformers, TCNs maintain a simpler structure while retaining parallelism and causality.

\section{Method} \label{sec:method}

\begin{figure*}[!t]
    \includegraphics[width=1.1\linewidth]{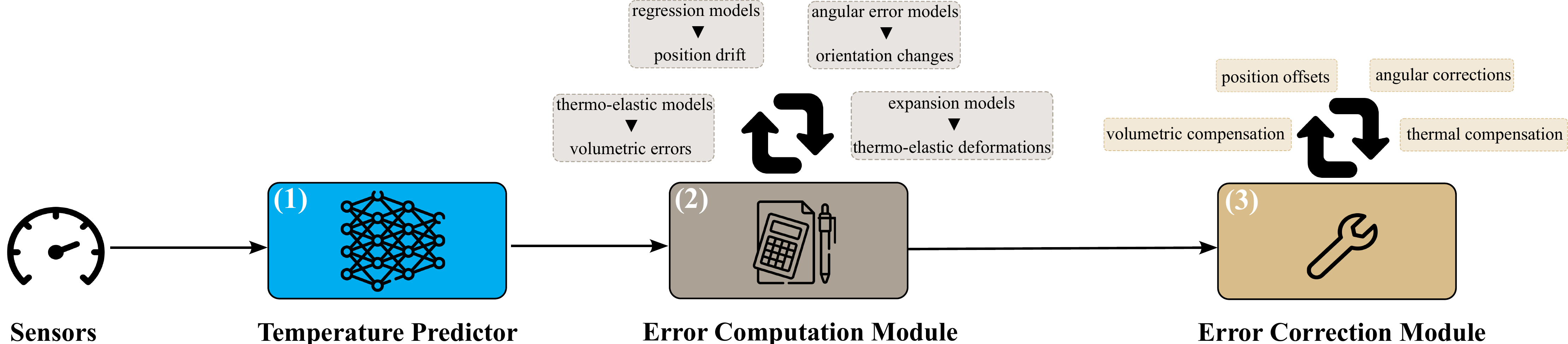}
    \caption{Schematic representation of the proposed thermal error compensation framework, consisting of three modules: (1) a temperature predictor to forecast temperature distributions given a known initial thermal field; (2) a swappable error computation module, which employs different strategies to compute various errors; (3) a swappable error correction module, which, given the error, computes the corresponding compensation.}
    \label{fig:framework}
\end{figure*}

In this work, we propose directly modelling the thermal field and heat flux of machine tools using NNs thus giving arise of a novel modular framework for error compensation, Figure \ref{fig:framework}.
It is composed of three main components (1) a temperature predictor, given by a NN, a swappable error computation module, and an error correction module:

\begin{enumerate}
    \item \emph{Temperature predictor.} A NN model that forecasts the thermal and heat flux distribution over time within the machine tool. To do this, the temperature in a known time step is given as input and predictions are made for subsequent time steps, seamlessly capturing both transient and steady-state phases of thermal evolution. The NN model enables real-time, high-resolution, and low cost thermal field predictions while reducing the dependence on dense sensor measurements and costly FEM simulations. Also, the NN model can generalise to previously unseen operational conditions since it learns thermodynamic phenomena governed by universal laws;

    \item \emph{Error computation module.} Given predictions of the thermal field and/or heat flux, this swappable module can use different error computation strategies based on physics-driven analytical models (relating thermal expansion to positional error), empirical regressions (extracting statistical relations between temperature and error), or lookup tables (rapid compensation in repetitive scenarios). This flexibility allows for the usage of different modules, able to compute a wide range of errors (such as volumetric errors, position drift, orientation changes, and thermo-elastic deformations), without modifying the overall framework;

    \item \emph{Error correction module.} Given the errors computed by the previous module, it computes the appropriate compensation to the machine tool based on the desired goal. Depending on the target it may compute optimal offset values for tool paths to correct positional drift, calculate angular correction parameters for spindle alignment, or construct error maps for controller integration. This swappable module allows flexibility to define compensation criteria and output, ensuring compatibility with various machine control systems, easing the implementation of different software and hardware compensation strategies.
\end{enumerate}

For the temperature predictor, we propose using time-series capable NNs, namely RNN, GRU, LSTM, BiLSTM, Transformer, and TCN. 
To train these NN architectures, the necessary data was generated using a simulation-based approach. First, a comprehensive Computer-Aided Design (CAD) model of a vertical machine tool was constructed through Solidworks. This CAD model was then streamlined through defeaturing and segmentation to retain essential geometric and thermal characteristics while optimising for computational efficiency. The refined CAD assembly was then used as input to a FEM thermal analysis conducted in ANSYS, where the thermal field and heat flux distributions were computed under different operating conditions and environmental scenarios. This process produced high-fidelity spatio-temporal temperature datasets that capture transient and steady-state thermal behaviours \cite{coelho2025open}. 

High-fidelity FEM simulations are computationally demanding, especially as the number of nodes increases to capture more detailed spatial and temporal temperature variations within the machine tool. Fine discretisation improves model accuracy, however raises computational costs and simulation time which can become prohibitive \cite{sachdeva2006comparative}. 
To address this challenge, we propose a two-stage node selection strategy: first, a correlation matrix analysis, on the FEM obtained data, is performed to assess the redundancy and dependency among node thermal fields and heat fluxes, thus allowing to disregard nodes with low information contribution; second, at inference, the NN predictions at the disregarded nodes are efficiently reconstructed using the retained, most informative nodes. This approach effectively reduces the number of required physical sensors, minimises overfitting risks for the NN, and preserves the prediction accuracy of the thermal fields. 

\begin{definition}[Two-stage node selection strategy] \label{def:node}
    The correlation matrix is generated by computing the Pearson correlation, $\rho_{k \ell}$, between all node pairs $k, \ell \in \{1, \dots, d\}$ \cite{benesty2009pearson}. Then, the nodes whose absolute correlation exceeds $\tau$ are discarded from the training dataset and only one is kept. Let $\mathcal{S} = \{ \Gamma_1, \dots, \Gamma_J \} \subseteq \{1, \dots, d\}$ be the indices of the retained (parent) nodes, where $J$ is the total number of kept nodes, and let $\psi: \{1, \dots, d\} \to \mathcal{S}$ denote the parent mapping (such as $\psi(k) = \arg \max_{j \in \mathcal{S}| \rho_{k j}|}$). Then, for every discarded node $k \notin \mathcal{S}$ we require $|\rho_{k \psi(k)}| > \tau$. Training of the NN architectures is performed with the $J$ kept nodes $\{{node}_{\Gamma_j}\}_{j=1}^J$. 

Then, at inference, each discarded node $k \notin \mathcal{S}$, with parent $j=\psi(k)$, is reconstructed from its parent $node_{\Gamma_j}$ using a linear mapping:

\begin{equation}
node_{k} = m_{k j} \cdot {node}_{\Gamma_j} + b_{k j},
\end{equation}

 \noindent where,

\begin{equation}
m_{k j} = \rho_{k \Gamma_j}\dfrac{\sigma_k}{\sigma_{\Gamma_j}} \,\,\, \text{and} \,\,\, b_{k j} = \mu_k - m_{k j}\mu_{\Gamma_j},
\end{equation}

\noindent where $\rho_{k \Gamma_j}$ is the Pearson coefficient between node $k$ and parent $\Gamma_j$, $\sigma_k, \sigma_{\Gamma_j}$ the standard deviation (std) at $k$ and $\Gamma_j$ respectively, and $\mu_k, \mu_{\Gamma_j}$ the mean at $k$ and $\Gamma_j$ respectively \cite{hastie2009elements}.
\end{definition}

\begin{figure*}[!t]
    \includegraphics[width=1.1\linewidth]{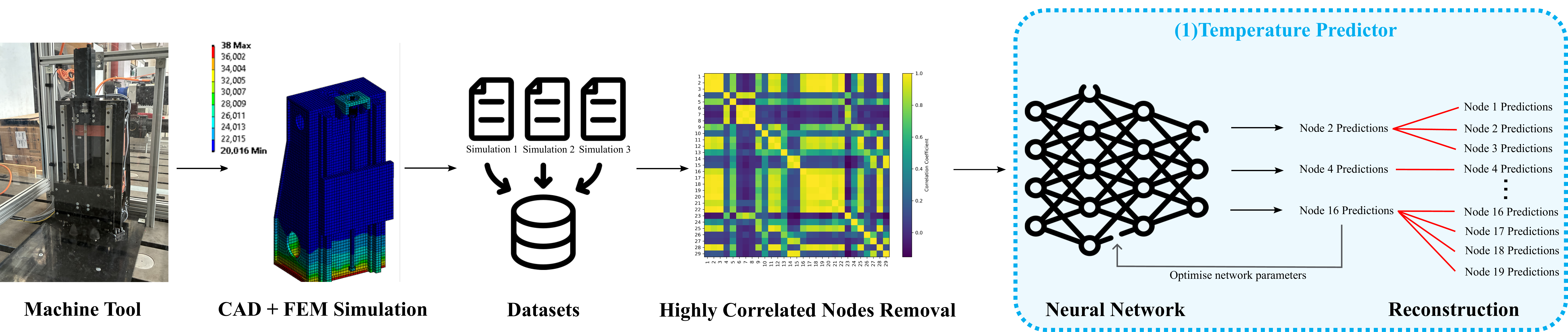}
    \caption{Schematic representation of the temperature predictor training pipeline, consisting of: a machine tool digital twin is constructed via CAD and FEM simulations to generate thermal field and heat flux datasets under different initial conditions; through correlation analysis, the lower informative nodes are removed for NN training, minimising prediction error on retained nodes; for inference, disregarded node values are reconstructed using retained nodes.}
    \label{fig:predictorPipeline}
\end{figure*}

Furthermore, we present two approaches for training the temperature predictor. First, a specialised predictor is trained using a single FEM simulation data conducted with one set of initial conditions (material density and thermal conductivity, and temperature at the starting time for all sensors, thus being tailored for extrapolation under fixed initial conditions. Second, a generalised predictor is trained using multiple FEM simulations data, each with different sets of initial conditions, allowing the NN to extrapolate thermal fields for previously unseen initial conditions.
The temperature predictor training pipeline is summarised in Figure \ref{fig:predictorPipeline}.

\section{Experimental Study} \label{sec:exps}

To validate our approach, we present experimental results of training and testing a temperature predictor using various NN architectures, namely RNN, GRU, LSTM, BiLSTM, Transformer, and TCN. Thus, also providing a benchmark for thermal field and heat flux modelling with NNs. The code implementations were done using \emph{Pytorch} and are accessible at [LINK]\footnote{The link to the repository will be made available after paper acceptance.}.




\subsection{Experimental Setup and Data Description}

In this work, an open accessible dataset collection was used \cite{fernandez_2025_17142176}. It provides a collection of datasets, from a vertical machine tool obtained with different initial conditions, with thermal field and heat flux values, from 29 sensors strategically positioned near critical heat flows (i.e. particularly at abrupt geometrical transitions and assembly interfaces) \cite{coelho2025open}. Using a large number of FEM nodes as direct inputs for the NN temperature predictor would imply that an equally large number of physical temperature sensors would be required on the machine to make predictions, which is impractical and costly.

The boundary conditions considered were environmental air temperature, ground temperature, initial system temperature, heat fluxes and air film coefficient. Heat fluxes were set as constant magnitudes, positioned on the motor support structure, bearings and guideways \cite{coelho2025open}. In our experiments we used 12 datasets, corresponding to the first 12 simulation runs. 

Note that this dataset collection does not account for certain non-linearities during the data generation in the FE model that could impact the analysis of thermal issues in machine tools. Specifically, friction, classified as a key factor under contact non-linearities and heat generation sources, was not considered, nor were the temperature-dependent variations in material properties, such as thermal conductivity and expansion coefficients. However, heat transfer non-linearities and time-dependent heat build-up were included in the modelling approach to address their influence on the thermal behaviour \cite{coelho2025open}.


Two distinct experiments, specialised and generalised prediction of the temperature or heat flux field, were performed. First, the NNs were trained separately on each dataset to predict either the thermal or the heat flux field. Second, we used leave-one-dataset cross-validation and trained the NNs on the combined remaining datasets while evaluating on the held-out dataset. In both experiments, highly correlated nodes with a Pearson coefficient $|\rho_{k \psi(k)}| > 0.95$ are removed and later reconstructed as described in Definition \ref{def:node} Section \ref{sec:method}.

For the thermal field, most nodes in the system are strongly correlated, indicating a uniform temperature distribution. Only a few nodes positioned in the front and back structure show weaker correlation. Contrarily, in the heat flux field fewer nodes are strongly correlated. Nodes in the carrier show strong correlation with each other, while motor base nodes only correlate with the middle guideway, indicating a heterogenous heat transfer of different components or locations in the system \cite{coelho2025open}.

For each model and experiment, hyperparameter tuning was performed using Optuna \cite{akiba2019optuna} with 20 trials, the searched parameter ranges are shown in Table \ref{tab:optuna_params}. The best hyperparameter settings for each architecture and experiment can be found in \ref{sec:hyperparams}. 

\begin{table}
    \centering
    \caption{Hyperparameter search space for Optuna optimisation of RNN, GRU, LSTM, BiLSTM, Transformer, and TCN. All ranges are inclusive.}
    \begin{tabular}{@{}lcc@{}}
         \toprule
         \textbf{Hyperparameter} & \textbf{Search Range} & \textbf{Model} \\
         \midrule
         Learning Rate & $10^{-5}$ -- $10^{-2}$ & All \\
         Weight Decay & $10^{-6}$ -- $10^{-3}$ & All \\
         Dropout & 0.1 -- 0.5 & All \\
         Number of Layers & 1 -- 12  & All\\
         Hidden Units &  32, 64, 128, 256 & All \\
         Number of Attention Heads & 2 -- 4 & Transformer \\
         \bottomrule
    \end{tabular}
    \label{tab:optuna_params}
\end{table}

Each architecture, experiment and best hyperparameters combination is trained for 30 epochs using the Adam with decoupled weight decay (AdamW) optimiser \cite{loshchilov2017decoupled} with a sequence length 10 and batch size 32. 

\subsection{Results and Discussion}

The resulting models were evaluated on the test or held-out dataset using the average mean squared error (MSE) and respective std over three runs. Additionally, average performance metrics for each model across all datasets, specialised and generalise, are presented.
Highly linear correlated feature nodes ($|\rho_{k \psi(k)}|>0.95$) were removed prior to training and then reconstructed with linear fitting. The experimental results can be found in Table \ref{tab:results_normal_temp} and Table \ref{tab:results_normal_heat} for the specialised models, and in Table \ref{tab:results_gen_temp} and Table \ref{tab:results_gen_heat} for the generalised models. The best model at each test dataset is highlighted in bold.

\begin{figure*}[htbp]
        \centering
        \begin{subfigure}[b]{0.3\textwidth}
            \includegraphics[width=\textwidth]{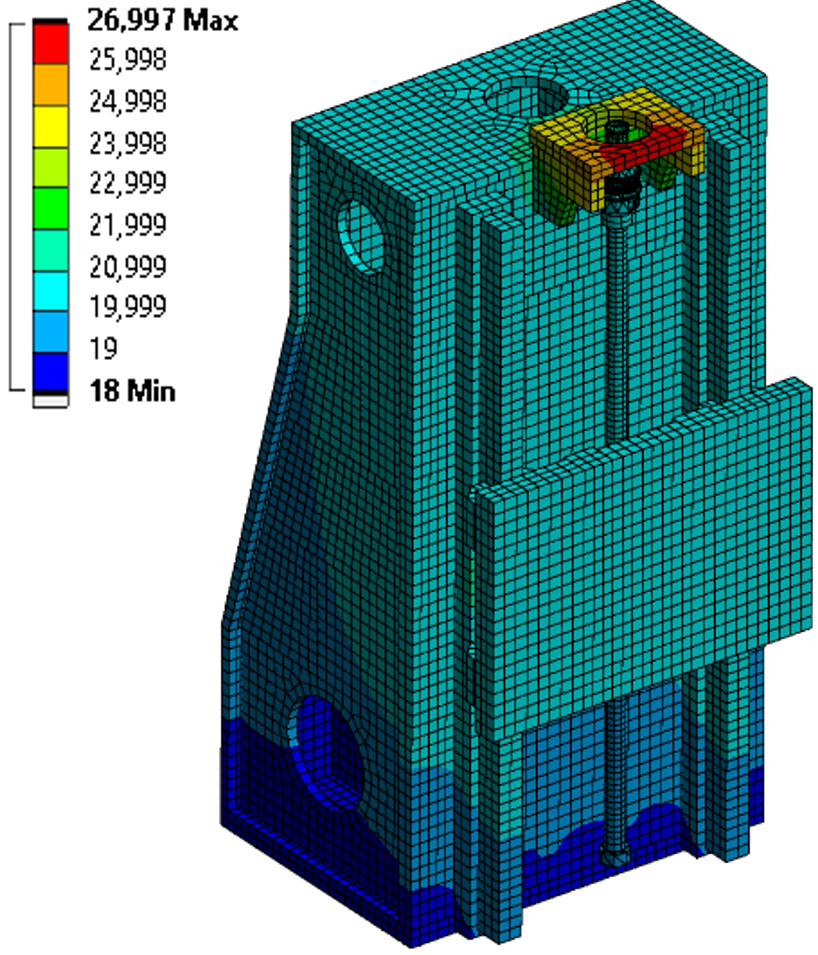}
            \caption{ANSYS Simulation}
        \end{subfigure}

        \begin{subfigure}[b]{0.3\textwidth}
            \includegraphics[width=\textwidth]{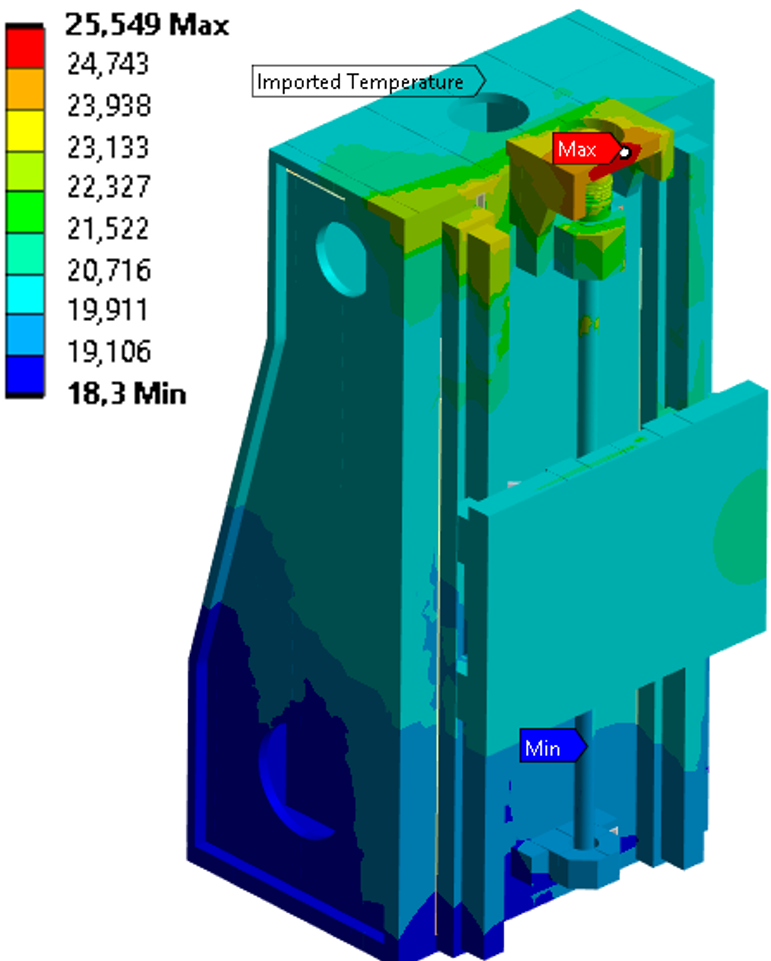}
            \caption{RNN}
        \end{subfigure}
        \begin{subfigure}[b]{0.3\textwidth}
            \includegraphics[width=\textwidth]{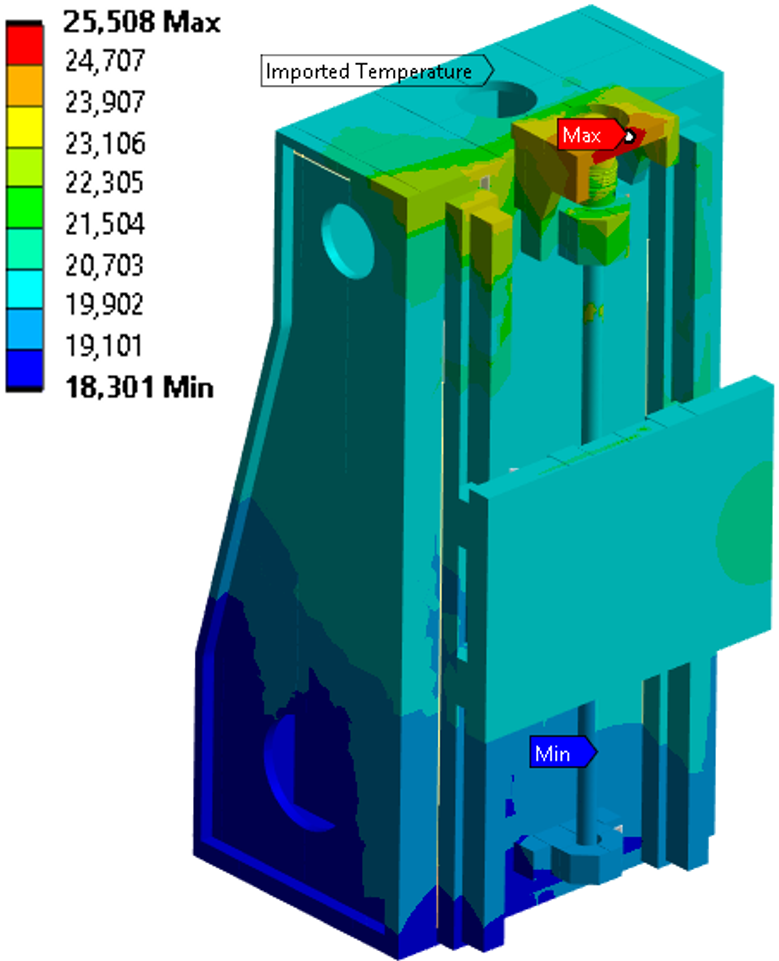}
            \caption{GRU}
        \end{subfigure}
        \begin{subfigure}[b]{0.3\textwidth}
            \includegraphics[width=\textwidth]{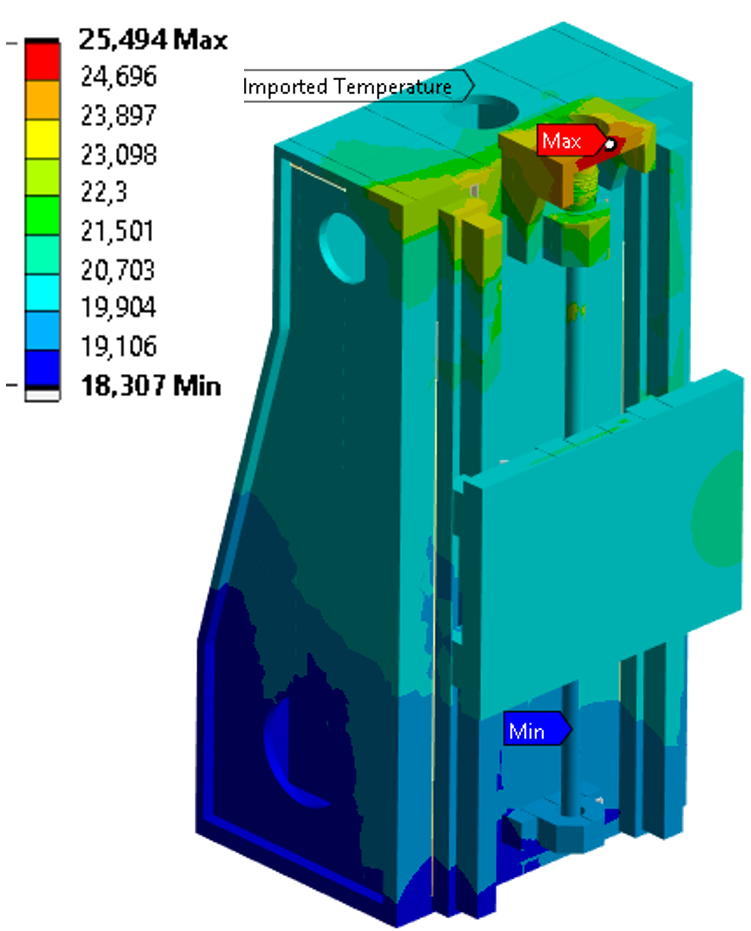}
            \caption{LSTM}
        \end{subfigure}
        
        \begin{subfigure}[b]{0.3\textwidth}
            \includegraphics[width=\textwidth]{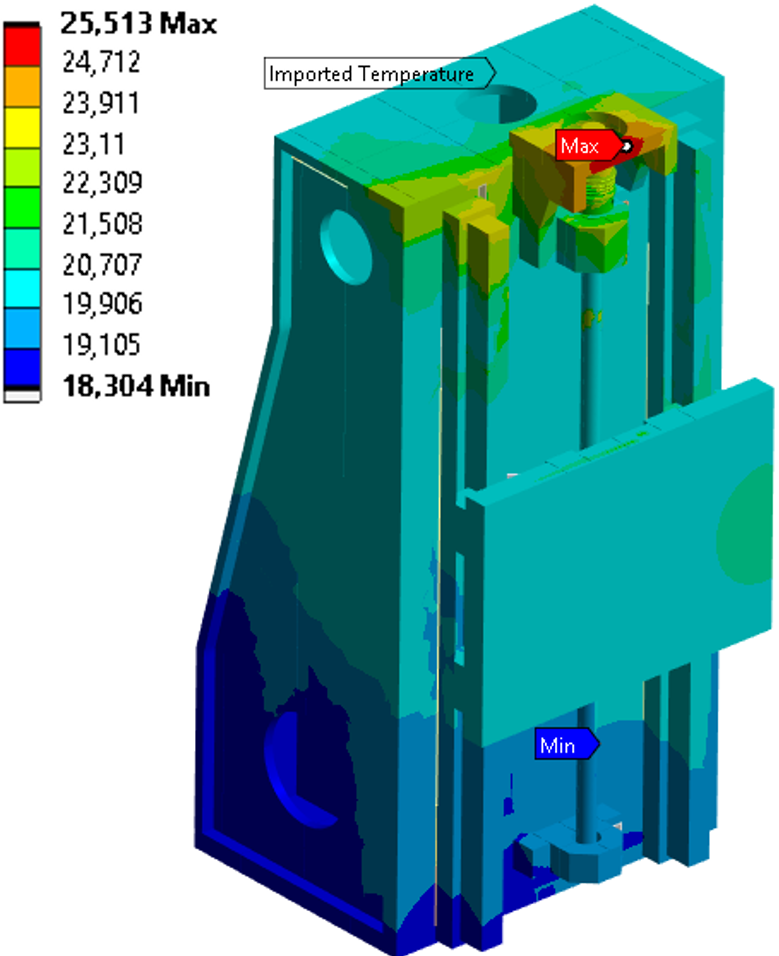}
            \caption{BiLSTM}
        \end{subfigure}
        \begin{subfigure}[b]{0.3\textwidth}

            \includegraphics[width=\textwidth]{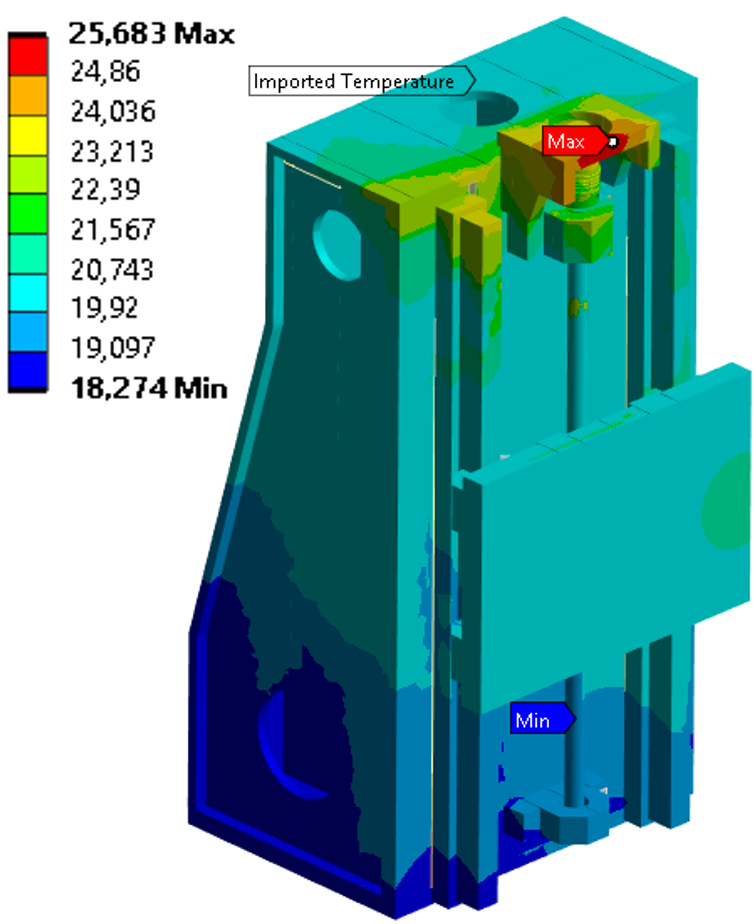}
            \caption{Transformer}
        \end{subfigure}
        \begin{subfigure}[b]{0.3\textwidth}
            \includegraphics[width=\textwidth]{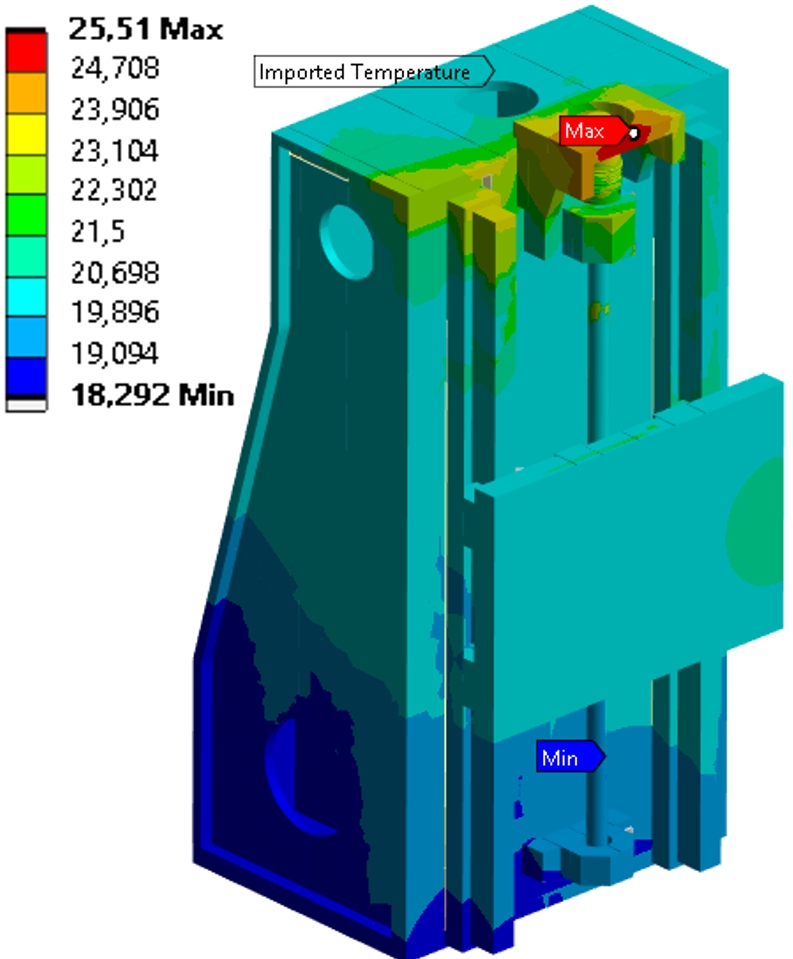}
            \caption{TCN}
        \end{subfigure}
    
        \caption{Comparison of thermal fields of RUN1 by interpolating temperatures from 29 sensors predicted the specialised temperature predictor with different NN architectures, and ANSYS simulation.}
        \label{fig:thermal-fields}
    \end{figure*}

\begin{figure*}[htbp]
        \centering
        \begin{subfigure}[b]{0.3\textwidth}
            \includegraphics[width=\textwidth]{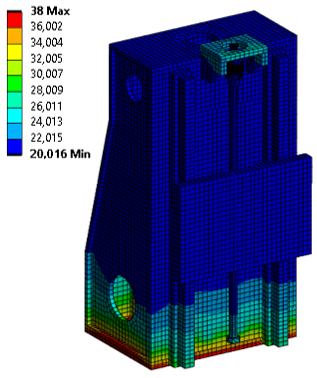}
            \caption{ANSYS Simulation}
        \end{subfigure}
            
        \begin{subfigure}[b]{0.3\textwidth}
            \includegraphics[width=\textwidth]{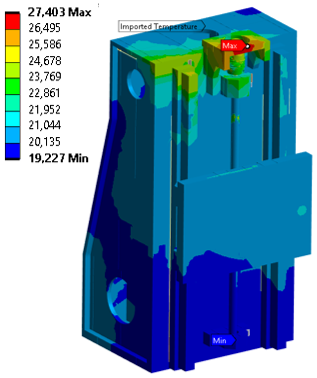}
            \caption{RNN}
        \end{subfigure}
        \begin{subfigure}[b]{0.3\textwidth}
            \includegraphics[width=\textwidth]{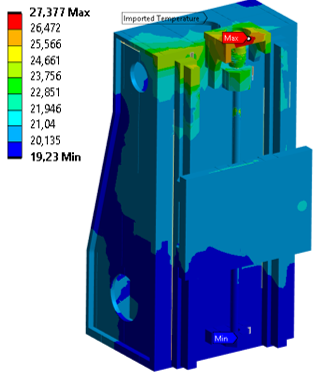}
            \caption{GRU}
        \end{subfigure}
        \begin{subfigure}[b]{0.3\textwidth}
            \includegraphics[width=\textwidth]{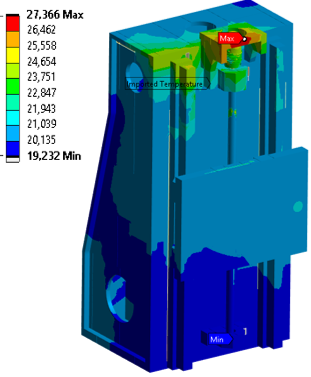}
            \caption{LSTM}
        \end{subfigure}        
        \begin{subfigure}[b]{0.3\textwidth}
            \includegraphics[width=\textwidth]{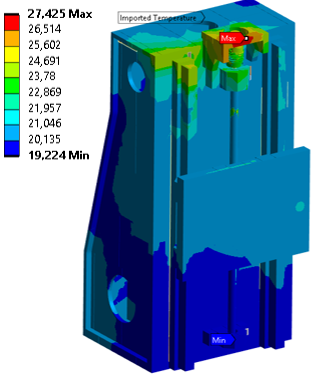}
            \caption{BiLSTM}
        \end{subfigure}
        \begin{subfigure}[b]{0.3\textwidth}
            \includegraphics[width=\textwidth]{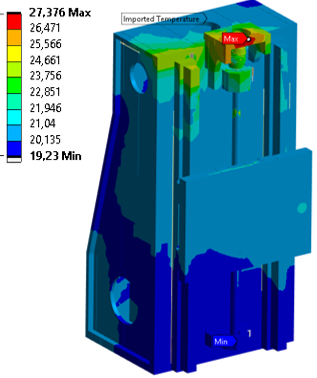}
            \caption{Transformer}
        \end{subfigure}
        \begin{subfigure}[b]{0.3\textwidth}
            \includegraphics[width=\textwidth]{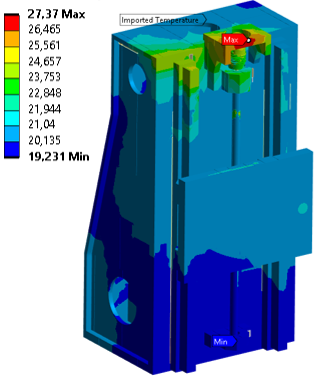}
            \caption{TCN}
        \end{subfigure}
    
        \caption{Comparison of thermal fields of Run3 by interpolating temperatures from 29 sensors predicted using different NN architectures, generalised approach and ANSYS simulation, also used as displayer.}
        \label{fig:thermal-fields2}
    \end{figure*}

Furthermore, to provide a direct and visual comparison between the thermal field predictions by the NN models and those produced by FEM simulation, the best run NNs, selected from the specialised (Table \ref{tab:results_normal_temp}) and generalised (Table \ref{tab:results_gen_temp}) predictor experiments, were evaluated at the final test time step. The predicted thermal field values, together with the sensor coordinates, were input into ANSYS Workbench for visualisation, as shown in Figure \ref{fig:thermal-fields} and Figure \ref{fig:thermal-fields2}. Given the relatively sparse sensor distribution compared to the high spatial resolution of the FEM simulation, linear interpolation was applied to estimate thermal fields between sensor locations. While this approach ensures basic continuity, it can compromise local accuracy and is sensitive to sensor placement. Note that, validation against direct ANSYS simulation (including explicit meshing and boundary conditions) confirmed the feasibility of this workflow, however it also demonstrated the need for denser sensor coverage in areas subject to high thermal gradients to enhance prediction quality.



\paragraph{Specialised Temperature Predictors}
\label{sec:specialised}

Table \ref{tab:results_normal_temp} and Table \ref{tab:results_normal_heat} present the results obtained for the NN architectures trained on thermal field and heat flux data generated by a single FEM simulation with one set of initial conditions. In total, 12 simulations with various initial conditions were performed, leading to 12 datasets for thermal field and 12 datasets for heat flux, that were modelled with RNN, GRU, LSTM, BiLSTM, Transformer, and TCN. Each dataset was split into $60\%$ training, validation $20\%$ and testing $20\%$.
The MSE and std are scaled by $\times 10^{-5}$. 

From Table \ref{tab:results_normal_temp}, GRU and BiLSTM achieve the best metrics on predicting the thermal field in most datasets. However, these architectures also produce remarkably high MSE and std variations in some datasets (RUN8, RUN11 and RUN12) leading to a degraded overall performance. TCN is the architecture that offers best metrics overall across all datasets while the Transformer is the worst.

From Table \ref{tab:results_normal_heat}, BiLSTM shows the best performance metrics for most datasets, 9 out of 12, while also having the best overall performance. In contrast, the Transformer and RNN display the worst performance.

Overall, GRU and BiLSTM are the best architectures for thermal field modelling with specialised temperature predictors.


\begin{table}
    \centering
    \caption{Results for thermal field prediction. Each model was trained and evaluated exclusively on one dataset. All values are scaled by $\times 10^{-5}$ and averaged over three runs. Input features with high linear correlation ($|\rho_{k \psi(k)}| > 0.95$) were removed.}
    \resizebox{\textwidth}{!}{%
\begin{tabular}{@{}c*{6}{c}@{}}
        \toprule
        \multirow{2}{*}{\textbf{Dataset}} & \textbf{RNN} & \textbf{GRU} & \textbf{LSTM} & \textbf{BiLSTM} & \textbf{Transformer} & \textbf{TCN} \\
        & MSE$\pm$STD & MSE$\pm$STD & MSE$\pm$STD & MSE$\pm$STD & MSE$\pm$STD & MSE$\pm$STD \\
        \midrule
            RUN1 & $11.57\pm5.80$ & $\mathbf{4.90}\pm0.10$ & $8.67\pm3.70$ & $5.13\pm0.78$ & $39.03\pm23.32$ & $5.70\pm1.39$ \\
            RUN2 & $3.13\pm0.92$ & $\mathbf{0.27}\pm0.21$ & $1.47\pm1.15$ & $0.33\pm0.42$ & $22.03\pm3.05$ & $1.20\pm1.56$ \\
            RUN3 & $4.37\pm3.00$ & $\mathbf{0.63}\pm0.40$ & $3.30\pm2.50$ & $0.80\pm0.78$ & $36.40\pm26.01$ & $0.90\pm0.95$ \\
            RUN4 & $11.23\pm6.25$ & $5.73\pm2.87$ & $5.67\pm0.80$ & $\mathbf{4.23}\pm1.43$ & $19.87\pm9.68$ & $4.80\pm0.56$ \\
            RUN5 & $2.13\pm0.71$ & $0.77\pm0.67$ & $0.83\pm0.59$ & $\mathbf{0.07}\pm0.12$ & $61.63\pm33.41$ & $0.27\pm0.15$ \\
            RUN6 & $0.50\pm0.26$ & $0.07\pm0.06$ & $0.10\pm0.17$ & $\mathbf{0.00}\pm0.00$ & $44.73\pm36.03$ & $\mathbf{0.00}\pm0.00$ \\
            RUN7 & $0.90\pm0.95$ & $0.03\pm0.06$ & $1.10\pm1.65$ & $\mathbf{0.00}\pm0.00$ & $45.57\pm25.37$ & $0.03\pm0.06$ \\
            RUN8 & $1.37\pm0.83$ & $17.50\pm30.31$ & $0.07\pm0.06$ & $0.17\pm0.29$ & $16.17\pm4.62$ & $\mathbf{0.03}\pm0.06$ \\
            RUN9 & $0.47\pm0.21$ & $\mathbf{0.00}\pm0.00$ & $0.20\pm0.17$ & $0.03\pm0.06$ & $20.83\pm3.50$ & $0.13\pm0.15$ \\
            RUN10 & $0.80\pm0.96$ & $\mathbf{0.13}\pm0.06$ & $0.53\pm0.23$ & $0.47\pm0.29$ & $18.50\pm14.97$ & $0.73\pm1.10$ \\
            RUN11 & $0.40\pm0.61$ & $0.47\pm0.72$ & $0.33\pm0.23$ & $\mathbf{0.07}\pm0.06$ & $7.00\pm4.00$ & $0.23\pm0.21$ \\
            RUN12 & $0.90\pm0.75$ & $0.33\pm0.25$ & $0.20\pm0.17$ & $8.63\pm14.95$ & $55.23\pm43.65$ & $\mathbf{0.13}\pm0.15$ \\
        \midrule
        Average & $3.15\pm1.77$ & $2.57\pm2.98$ & $1.87\pm0.95$ & $1.66\pm1.60$ & $32.25\pm18.97$ & $\mathbf{1.18}\pm0.53$ \\
        \bottomrule
        \end{tabular}%
    }
    \label{tab:results_normal_temp}
\end{table}

\begin{table}
    \centering
    \caption{Results for heat flux prediction. Each model was trained and evaluated exclusively on one dataset. All values are scaled by $\times 10^{-5}$ and averaged over three runs. Input features with high linear correlation ($|\rho_{k \psi(k)}| > 0.95$) were removed.}
    \resizebox{\textwidth}{!}{%
\begin{tabular}{@{}c*{6}{c}@{}}
       \toprule
        \multirow{2}{*}{\textbf{Dataset}} & \textbf{RNN} & \textbf{GRU} & \textbf{LSTM} & \textbf{BiLSTM} & \textbf{Transformer} & \textbf{TCN} \\
        & MSE$\pm$STD & MSE$\pm$STD & MSE$\pm$STD & MSE$\pm$STD & MSE$\pm$STD & MSE$\pm$STD \\
        \midrule
            RUN1 & $6.63\pm6.92$ & $\mathbf{0.40}\pm0.10$ & $1.43\pm0.45$ & $\mathbf{0.40}\pm0.26$ & $10.63\pm4.44$ & $0.77\pm0.47$ \\
            RUN2 & $2.77\pm0.83$ & $0.73\pm0.21$ & $2.20\pm1.54$ & $\mathbf{0.27}\pm0.21$ & $14.60\pm9.18$ & $1.47\pm1.01$ \\
            RUN3 & $7.07\pm5.51$ & $0.37\pm0.25$ & $1.10\pm0.62$ & $\mathbf{0.13}\pm0.06$ & $5.93\pm2.97$ & $0.50\pm0.17$ \\
            RUN4 & $4.93\pm1.95$ & $1.03\pm0.59$ & $1.80\pm0.36$ & $\mathbf{0.50}\pm0.61$ & $9.60\pm1.83$ & $1.03\pm0.38$ \\
            RUN5 & $7.23\pm5.90$ & $\mathbf{1.07}\pm0.45$ & $2.57\pm0.47$ & $1.30\pm1.56$ & $16.73\pm11.95$ & $1.43\pm0.76$ \\
            RUN6 & $13.00\pm10.86$ & $2.23\pm1.63$ & $2.93\pm1.04$ & $\mathbf{0.43}\pm0.32$ & $9.23\pm3.35$ & $1.33\pm0.55$ \\
            RUN7 & $6.67\pm4.55$ & $0.80\pm0.26$ & $1.27\pm0.32$ & $\mathbf{0.43}\pm0.21$ & $16.57\pm7.21$ & $0.87\pm0.47$ \\
            RUN8 & $3.13\pm0.78$ & $0.47\pm0.06$ & $1.47\pm0.15$ & $0.70\pm0.44$ & $9.33\pm4.76$ & $\mathbf{0.43}\pm0.32$ \\
            RUN9 & $9.83\pm6.03$ & $1.27\pm0.76$ & $5.27\pm2.91$ & $\mathbf{0.57}\pm0.55$ & $18.07\pm12.25$ & $1.73\pm1.01$ \\
            RUN10 & $4.17\pm1.27$ & $0.53\pm0.06$ & $1.80\pm0.56$ & $\mathbf{0.37}\pm0.23$ & $9.40\pm2.55$ & $0.87\pm0.25$ \\
            RUN11 & $3.73\pm3.87$ & $0.43\pm0.32$ & $1.97\pm0.95$ & $\mathbf{0.30}\pm0.35$ & $14.20\pm9.35$ & $0.77\pm0.45$ \\
            RUN12 & $6.77\pm4.48$ & $4.77\pm6.80$ & $4.33\pm5.95$ & $1.43\pm1.80$ & $13.50\pm1.66$ & $\mathbf{0.67}\pm0.06$ \\
        \midrule
        Average & $6.33\pm4.41$ & $1.18\pm0.96$ & $2.35\pm1.28$ & $\mathbf{0.57}\pm0.55$ & $12.32\pm5.96$ & $0.99\pm0.49$ \\
        \bottomrule
        \end{tabular}%
    }
    \label{tab:results_normal_heat}
\end{table}

Figure \ref{fig:thermal-fields} shows a comparative visualisation of the thermal field predictions with a specialised temperature predictor for RUN1 using the best-performing runs by RNN, GRU, LSTM, BiLSTM, Transformer, and TCN, alongside the reference FEM simulation in ANSYS. All NN models successfully capture the essential features of the thermal distribution, accurately identifying both the locations and magnitudes of highest and lowest temperature regions. Furthermore, the predicted temperature ranges closely match those of the FEM simulation. This consistency across architectures shows the robustness of NN-based approaches in reproducing thermal behaviours in machine tools, despite differences in model structure.

\paragraph{Generalised Temperature Predictors}

Table \ref{tab:results_gen_temp} and Table \ref{tab:results_gen_heat} present the results obtained for the NN architectures trained on thermal field and heat flux data generated by 12 FEM simulations with various sets of initial conditions. To ensure robust evaluation and fairness, we employed a leave-one-out cross-validation strategy with each of the 12 simulations held out in turn as a test set. 
The MSE and std are scaled by $\times 10^{-5}$. 

From Table \ref{tab:results_gen_temp}, GRU offers, in general, the best performance metrics in all held-out datasets (8 out of 12) and overall. In comparison with the specialised temperature predictors (Table \ref{tab:results_normal_temp} and Table \ref{tab:results_normal_heat}), the Transformer architecture performs better in the generalised case, likely due to the larger amount of training data available. 
As expected, RNN continues to offer the worst overall performance.

From Table \ref{tab:results_gen_heat}, once again GRU displays the best overall performance out of all architectures. However, all models show a one order magnitude increase of MSE, showing generalising heat flux dynamics proves to be harder. Particularly, for the held-out datasets RUN7, RUN8 and RUN11, all architectures perform poorly, explained due to the unique initial conditions that makes the heat flux behaviour be very different form the other datasets. It is important to note that the Transformer bad overall performance is due to RUN7, where the MSE is $\approx 4$ times higher than the second worst performant architecture, LSTM.

Overall, GRU is the best architecture for heat flux modelling with generalised temperature predictors.

\begin{table}
    \centering
    \caption{Results for thermal field prediction. Models were trained using leave-one-out cross-validation, i.e. each model is trained on the combined remaining datasets, and evaluated on the held-out dataset. All values are scaled by $\times 10^{-5}$ and averaged over three runs. Input features with high linear correlation ($|\rho_{k \psi(k)}| > 0.95$) were removed.}
    \resizebox{\textwidth}{!}{%
 \begin{tabular}{@{}c*{6}{c}@{}}
        \toprule
        \multirow{2}{*}{\textbf{Held-Out}} & \textbf{RNN} & \textbf{GRU} & \textbf{LSTM} & \textbf{BiLSTM} & \textbf{Transformer} & \textbf{TCN} \\
        & MSE$\pm$STD & MSE$\pm$STD & MSE$\pm$STD & MSE$\pm$STD & MSE$\pm$STD & MSE$\pm$STD \\
        \midrule
            RUN1 & $4.17\pm1.70$ & $\mathbf{1.10}\pm0.36$ & $1.70\pm0.17$ & $1.97\pm1.55$ & $3.40\pm1.59$ & $1.60\pm1.15$ \\
            RUN2 & $29.23\pm17.20$ & $6.17\pm2.35$ & $6.40\pm4.03$ & $5.67\pm2.15$ & $11.60\pm5.37$ & $\mathbf{5.03}\pm4.67$ \\
            RUN3 & $27.63\pm12.38$ & $8.57\pm5.76$ & $18.23\pm6.91$ & $11.53\pm7.25$ & $\mathbf{6.70}\pm1.25$ & $18.03\pm10.22$ \\
            RUN4 & $12.83\pm8.21$ & $\mathbf{2.30}\pm1.65$ & $8.60\pm8.96$ & $4.83\pm1.83$ & $8.27\pm2.08$ & $2.93\pm0.68$ \\
            RUN5 & $18.83\pm15.45$ & $\mathbf{1.57}\pm0.40$ & $1.90\pm1.04$ & $10.83\pm6.50$ & $9.80\pm3.95$ & $9.53\pm8.62$ \\
            RUN6 & $32.20\pm27.50$ & $\mathbf{1.97}\pm0.60$ & $7.53\pm2.75$ & $2.47\pm1.42$ & $4.00\pm0.78$ & $4.10\pm1.05$ \\
            RUN7 & $4.43\pm1.85$ & $\mathbf{2.03}\pm0.23$ & $3.27\pm0.60$ & $6.97\pm7.66$ & $3.40\pm2.29$ & $2.80\pm1.13$ \\
            RUN8 & $11.27\pm2.95$ & $2.60\pm0.96$ & $\mathbf{2.00}\pm1.39$ & $3.63\pm0.96$ & $9.10\pm5.74$ & $2.30\pm2.17$ \\
            RUN9 & $3.97\pm0.90$ & $\mathbf{1.37}\pm0.40$ & $2.13\pm1.53$ & $2.17\pm0.45$ & $3.87\pm1.74$ & $1.93\pm1.18$ \\
            RUN10 & $9.67\pm6.70$ & $2.90\pm1.67$ & $8.73\pm8.56$ & $3.73\pm0.40$ & $5.90\pm3.70$ & $\mathbf{1.40}\pm0.20$ \\
            RUN11 & $3.10\pm0.79$ & $\mathbf{2.43}\pm0.50$ & $\mathbf{2.43}\pm1.54$ & $5.77\pm1.88$ & $5.03\pm3.71$ & $3.83\pm3.44$ \\
            RUN12 & $21.87\pm14.94$ & $\mathbf{1.57}\pm0.57$ & $4.30\pm0.78$ & $6.40\pm5.35$ & $7.97\pm6.74$ & $2.83\pm2.27$ \\
        \midrule
        Average & $14.93\pm9.21$ & $\mathbf{2.88}\pm1.29$ & $5.60\pm3.19$ & $5.50\pm3.12$ & $6.59\pm3.25$ & $4.69\pm3.06$ \\
        \bottomrule
        \end{tabular}%
    }
    \label{tab:results_gen_temp}
\end{table}

\begin{table}
    \centering
    \caption{Results for heat flux prediction. Models were trained using leave-one-out cross-validation, i.e. each model is trained on the combined remaining datasets, and evaluated on the held-out dataset. All values are scaled by $\times 10^{-5}$ and averaged over three runs. Input features with high linear correlation ($|\rho_{k \psi(k)}| > 0.95$) were removed.}
    \resizebox{\textwidth}{!}{%
\begin{tabular}{@{}c*{6}{c}@{}}
        \toprule
        \multirow{2}{*}{\textbf{Held-Out}} & \textbf{RNN} & \textbf{GRU} & \textbf{LSTM} & \textbf{BiLSTM} & \textbf{Transformer} & \textbf{TCN} \\
        & MSE$\pm$STD & MSE$\pm$STD & MSE$\pm$STD & MSE$\pm$STD & MSE$\pm$STD & MSE$\pm$STD \\
        \midrule
            RUN1 & $3.67\pm0.51$ & $\mathbf{0.96}\pm0.25$ & $2.03\pm0.85$ & $2.36\pm1.09$ & $2.27\pm0.32$ & $1.13\pm0.85$ \\
            RUN2 & $15.27\pm9.87$ & $2.80\pm1.21$ & $6.33\pm4.80$ & $20.43\pm11.74$ & $15.87\pm18.32$ & $\mathbf{2.63}\pm1.59$ \\
            RUN3 & $7.70\pm3.12$ & $3.03\pm0.60$ & $8.00\pm3.70$ & $19.27\pm9.16$ & $15.40\pm5.35$ & $\mathbf{2.40}\pm1.56$ \\
            RUN4 & $21.20\pm7.04$ & $\mathbf{3.83}\pm0.85$ & $14.17\pm2.76$ & $8.40\pm1.04$ & $24.83\pm21.43$ & $37.03\pm21.89$ \\
            RUN5 & $34.90\pm9.11$ & $\mathbf{8.07}\pm1.76$ & $32.23\pm21.04$ & $72.43\pm50.30$ & $24.53\pm11.83$ & $14.37\pm11.69$ \\
            RUN6 & $11.70\pm6.94$ & $12.60\pm6.07$ & $32.27\pm9.29$ & $29.63\pm11.66$ & $\mathbf{9.66}\pm1.79$ & $40.13\pm63.91$ \\
            RUN7 & $165.67\pm127.14$ & $\mathbf{142.13}\pm55.81$ & $217.37\pm231.16$ & $177.47\pm19.91$ & $899.00\pm688.94$ & $193.80\pm168.56$ \\
            RUN8 & $132.43\pm85.38$ & $87.67\pm16.73$ & $188.37\pm97.44$ & $82.50\pm35.30$ & $190.83\pm62.67$ & $\mathbf{65.93}\pm72.03$ \\
            RUN9 & $44.50\pm24.29$ & $25.13\pm4.07$ & $55.07\pm16.44$ & $39.93\pm7.22$ & $\mathbf{20.20}\pm5.22$ & $36.40\pm17.59$ \\
            RUN10 & $42.40\pm45.76$ & $23.90\pm6.10$ & $42.50\pm36.42$ & $47.97\pm18.60$ & $\mathbf{17.27}\pm4.07$ & $24.37\pm30.69$ \\
            RUN11 & $\mathbf{64.17}\pm37.54$ & $82.57\pm57.79$ & $173.17\pm22.80$ & $82.30\pm38.29$ & $160.27\pm64.38$ & $128.70\pm99.16$ \\
            RUN12 & $28.93\pm9.73$ & $\mathbf{8.80}\pm2.78$ & $34.87\pm3.11$ & $34.23\pm12.77$ & $44.50\pm20.04$ & $9.73\pm7.45$ \\
        \midrule
        Average & $47.71\pm30.54$ & $\mathbf{33.46}\pm12.84$ & $67.20\pm37.48$ & $51.41\pm18.09$ & $118.72\pm75.36$ & $46.38\pm41.41$ \\
        \bottomrule
        \end{tabular}%
    }
    \label{tab:results_gen_heat}
\end{table}



In general, across both experiments, GRU shows to offer the best all-around performance. The results obtained, for the specialised and generalised temperature predictors, are in accordance with literature. 

The simplicity of RNN hinders its ability to learn long-term dependencies being overpassed by GRU and LSTM. Although LSTM has a more complex gating mechanism than GRU, allowing for learning more complex behaviours, the increase in the number of parameters increases the optimisation process difficulty. BiLSTM shows remarkably better performance than LSTM which is can be attributed to the usage of the training dataset forwards and backwards in time which effectively augmentates the amount of data available \cite{coelho2025neural}. The Transformer shows the worst performance which can be attributed to the high number of parameters of the architecture that would require larger amounts of data and more training epochs. In contrast, TCN effectively reduces the amount of parameters, when compared with Transformers, while avoiding recurrence.

Figure \ref{fig:thermal-fields2} shows a comparative visualisation of the thermal field predictions with a generalised temperature predictor tested with RUN3 using the best-performing runs by RNN, GRU, LSTM, BiLSTM, Transformer, and TCN, alongside the reference FEM simulation in ANSYS.
As expected, the thermal fields predicted by the generalised NN models do not match the FEM simulation as closely as those generated by the specialised temperature predictors, Figure \ref{fig:thermal-fields2}. This outcome reflects the increased challenge posed by extrapolating to unseen initial conditions, which presents greater variability and complexity for the NNs. Nonetheless, each architecture accurately identified key distribution features in which the upper region of the machine consistently exhibited elevated temperatures, in alignment with the FEM results, while the minimum recorded temperatures were also in close agreement. These findings highlight the strengths and limitations of the generalised frameworks, hinting that expanding the diversity and quantity of the datasets, with different initial conditions, is critical for further improving the prediction performance and generalisation ability of these models.

\clearpage
\section{Conclusion} \label{sec:conc}

In this work, we establish a novel paradigm for thermal error compensation in machine tools by using neural networks to predict thermal field and heat flux, moving beyond existing literature approaches that directly model thermal error or compensation. Decoupling thermal state from error computation allows the temperature predictor to be flexibly reused for the computation and compensation of various types of thermal errors across different machine configurations. The proposed an accompanying framework comprises of three components: a neural network-based temperature predictor; an error computation module; an error correction module. 

The temperature predictor is trained using FEM-generated datasets under multiple initial conditions, and supplemented by a correlation-based node selection strategy to identify the most informative nodes, thus minimising sensor requirements during inference. Also, a reconstruction approach leveraging the predicted nodes and using linear regression is proposed.

A thorough benchmark of state-of-the-art time-series capable neural network architectures (Recurrent Neural Network, Gated Recurrent Unit, Long-Short Term Memory, Bidirectional Long-Short Term Memory, Transformer, and Temporal Convolutional Network), considered both specialised models (trained on a single set of initial conditions) and generalised models (trained on multiple scenarios).

Experimental results show that both specialised and generalised models deliver high-quality predictions of thermal field and heat flux, with Gated Recurrent Unit architectures demonstrating the best overall performance. Notably, modelling heat flux dynamics proved to be more complex for generalised models, showing one order of magnitude higher error than specialised models. 
Visualisations of the NN-predicted fields confirm their ability to identify the location and magnitude of temperature extrema, with close alignment to FEM simulations. However, the prediction performance of generalised models highlights the importance of dataset diversity and sensor placement, as interpolation accuracy and thermal field resolution are sensitive to these factors. 

These findings emphasise that expanding dataset coverage and refining interpolation strategies are essential steps for enhancing robustness, accuracy, and generalisation capability of the NN models in machine tools




\section*{Acknowledgment}

C. Coelho would like to thank the KIBIDZ project funded by dtec.bw---Digitalization and Technology Research Center of the Bundeswehr; dtec.bw is funded by the European Union---NextGenerationEU.
M. Hohmann is funded by the Deutsche Forschungsgemeinschaft (DFG, German Research Foundation) – 520460697.
D. Fernández is funded by the Deutsche Forschungsgemeinschaft (DFG, German Research Foundation)  - F-009765-551-521-1130701.

\clearpage
\bibliography{00lib}
\bibliographystyle{IEEEtran}

\appendix

\sisetup{round-mode=figures, round-precision=3, scientific-notation=true}

\newpage

\section{Additional Hyperparameter Information}
\label{sec:hyperparams}

Here, additional information on the best parameters, for each architecture, found with Optuna for each dataset and hold-out dataset are presented. Tables \ref{tab:params_run1}-\ref{tab:params_run12} refer to specialised temperature predictors while Tables \ref{tab:gen_params_run1}-\ref{tab:gen_params_run12}.

\subsection{Specialised Temperature Predictors}

\begin{table}[h!]
    \centering
    \caption{Best hyperparameters for dataset RUN1.}
    \resizebox{\textwidth}{!}{%
        \begin{tabular}{@{}llccccccc@{}}
        \toprule
        Target & Model & Learning Rate & Weight Decay & Dropout  & Hidden Units & Layers & Att. Heads \\
        \midrule
        Temperature  & GRU          & \num{0.004526111285894689} & \num{0.0001806419802679935} & \num{0.2972042438671855} & 256 & 1 & - \\
        Temperature  & RNN          & \num{0.00022459451940676686}& \num{6.167008318244116e-05} & \num{0.18836518774568986}& 256 & 8 & - \\
        Temperature  & LSTM         & \num{0.0021285219101806526}& \num{0.0005863486355383001}  & \num{0.4973273987757159} & 256 & 1 & - \\
        Temperature  & BiLSTM       & \num{0.006187417764514722} & \num{0.0009678267164753225}  & \num{0.20872973500547337}& 256 & 2 & - \\
        Temperature  & TCN          & \num{0.006626854114135719} & \num{5.9154778115136434e-05} & \num{0.48206999841547865}&  64 & 1 & - \\
        Temperature  & Transformer  & \num{3.2491749523958704e-05}& \num{1.1500316115738665e-06} & \num{0.10247901104058174}&  32 & 3 & 4 \\
        Heat flux    & GRU          & \num{0.004795441307765604} & \num{0.0003183220262821532} & \num{0.4381947331800928} & 64  & 1 & - \\
        Heat flux    & RNN          & \num{0.001362074961919873} & \num{0.00039592053137795576}& \num{0.26334471706992607}& 256 & 7 & - \\
        Heat flux    & LSTM         & \num{0.00953416156261577}  & \num{0.000928519843472704}   & \num{0.17232853694631747}& 32  & 1 & - \\
        Heat flux    & BiLSTM       & \num{0.005555696008673844} & \num{0.0009356008187247625}  & \num{0.15484784977985905}& 128 & 1 & - \\
        Heat flux    & TCN          & \num{0.0022775359305039927}& \num{0.00011876079730592756} & \num{0.10056658310552635}& 256 & 1 & - \\
        Heat flux    & Transformer  & \num{0.0019170542963092149}& \num{0.00026113097884195523} & \num{0.26534446661041566}& 128 & 1 & 4 \\
        \bottomrule
        \end{tabular}%
    }
    \label{tab:params_run1}
\end{table}

\begin{table}[h!]
    \centering
    \caption{Best hyperparameters for dataset RUN2.}
    \resizebox{\textwidth}{!}{%
    \begin{tabular}{@{}llccccccc@{}}
        \toprule
        Target & Model & Learning Rate & Weight Decay & Dropout  & Hidden Units & Layers & Att. Heads \\
        \midrule
        Temperature  & GRU          & \num{0.0028159880133453367}& \num{1.3388140817120888e-05} & \num{0.38591318681206493}& 256 & 1 & - \\
        Temperature  & RNN          & \num{0.00048722645432565414}& \num{0.0009340127998612531}  & \num{0.34671452203513525}& 256 & 1 & - \\
        Temperature  & LSTM         & \num{0.0041413806279077935}& \num{1.172702598701168e-06}  & \num{0.3381083891855824} &  64 & 1 & - \\
        Temperature  & BiLSTM       & \num{0.0031168470566838814}& \num{1.3559055864149563e-05} & \num{0.13929353629434893}&  64 & 2 & - \\
        Temperature  & TCN          & \num{0.005563500462337619} & \num{0.00020031393440762024} & \num{0.24350103432191766}&  64 & 8 & - \\
        Temperature  & Transformer  & \num{2.7773471068926196e-05}& \num{1.0887607311221028e-06} & \num{0.10100445401036254}&  64 & 5 & 2 \\
        Heat flux    & GRU          & \num{0.001608454737332321} & \num{3.0775834069242164e-06} & \num{0.1608186153943058} & 256 & 5 & - \\
        Heat flux    & RNN          & \num{0.009297453034344887} & \num{6.358692919986704e-05}  & \num{0.17309433531169752}& 128 & 1 & - \\
        Heat flux    & LSTM         & \num{0.0018761381195798255}& \num{4.441888593145438e-06}  & \num{0.3823420340603947} & 256 & 1 & - \\
        Heat flux    & BiLSTM       & \num{0.0033426776699258267}& \num{0.000849568659495467}   & \num{0.11733913589238204}& 256 & 1 & - \\
        Heat flux    & TCN          & \num{0.001792377282896054} & \num{2.955577851539139e-05}  & \num{0.14328362592567173}& 128 & 1 & - \\
        Heat flux    & Transformer  & \num{7.575078675554828e-05} & \num{3.708197484872458e-05}  & \num{0.10023788607521479}&  64 & 6 & 2 \\
        \bottomrule
    \end{tabular}%
    }
    \label{tab:params_run2}
\end{table}

\begin{table}[h!]
    \centering
    \caption{Best hyperparameters for dataset RUN3.}
    \resizebox{\textwidth}{!}{%
    \begin{tabular}{@{}llccccccc@{}}
        \toprule
        Target & Model & Learning Rate & Weight Decay & Dropout  & Hidden Units & Layers & Att. Heads \\
        \midrule
        Temperature  & GRU          & \num{0.0005569431176264982}& \num{3.0424699310865583e-06} & \num{0.18663307438805324}& 256 & 2 & - \\
        Temperature  & RNN          & \num{0.0004260747912620301}& \num{0.00010093944664063393} & \num{0.2248311734949582} & 256 & 1 & - \\
        Temperature  & LSTM         & \num{0.002323879061554039} & \num{0.00040111859335996403} & \num{0.20798531948092497}& 256 & 4 & - \\
        Temperature  & BiLSTM       & \num{0.005460328801215009} & \num{0.00037538937696602996} & \num{0.3286174420186483} & 256 & 1 & - \\
        Temperature  & TCN          & \num{0.0014627658540974579}& \num{0.0009909499739545506}  & \num{0.22040712736030516}& 128 & 1 & - \\
        Temperature  & Transformer  & \num{0.00023486791909770594}& \num{1.2449574738690862e-06} & \num{0.3248373795801942} & 256 & 1 & 4 \\
        Heat flux    & GRU          & \num{0.004729743924405558} & \num{1.4380991898401532e-05} & \num{0.15335129670371472}& 256 & 1 & - \\
        Heat flux    & RNN          & \num{0.00971375524552094}  & \num{7.98343683641987e-06}   & \num{0.29479278190369135}&  64 & 1 & - \\
        Heat flux    & LSTM         & \num{0.0009357276360154747}& \num{7.153801156111398e-05}  & \num{0.24784150730589943}& 256 & 1 & - \\
        Heat flux    & BiLSTM       & \num{0.0015809650318409274}& \num{1.874871325220106e-05}  & \num{0.3429382224774211} & 256 & 3 & - \\
        Heat flux    & TCN          & \num{0.006080821699681034} & \num{2.039374194355209e-05}  & \num{0.49681614905325105}& 128 & 1 & - \\
        Heat flux    & Transformer  & \num{0.000548827359285584} & \num{4.125038592101575e-05}  & \num{0.41396031958881047}& 256 & 1 & 2 \\
        \bottomrule
    \end{tabular}%
    }
    \label{tab:params_run3}
\end{table}

\begin{table}[h!]
    \centering
    \caption{Best hyperparameters for dataset RUN4.}
    \resizebox{\textwidth}{!}{%
    \begin{tabular}{@{}llccccccc@{}}
        \toprule
        Target & Model & Learning Rate & Weight Decay & Dropout  & Hidden Units & Layers & Att. Heads \\
        \midrule
        Temperature  & GRU          & \num{0.002477697408007335} & \num{3.171824905968513e-05}  & \num{0.15762634609618725}&  64 & 1 & - \\
        Temperature  & RNN          & \num{0.0019229234267271853}& \num{0.000833718700069486}   & \num{0.10541587591788332}&  64 & 3 & - \\
        Temperature  & LSTM         & \num{0.009882599165965}    & \num{1.0051335542641648e-05} & \num{0.34116866785702854}& 128 & 1 & - \\
        Temperature  & BiLSTM       & \num{0.0024326147311211806}& \num{1.3966300183700262e-05} & \num{0.3904918681651218} & 128 & 1 & - \\
        Temperature  & TCN          & \num{0.007794183943575946} & \num{8.260680658973402e-05}  & \num{0.42435549482977375}&  32 & 2 & - \\
        Temperature  & Transformer  & \num{0.0004413252956543109}& \num{2.4969023169050884e-05} & \num{0.2146889471887381} & 128 & 2 & 4 \\
        Heat flux    & GRU          & \num{0.005167298390101216} & \num{0.00023701664236211976} & \num{0.3863601549733755} & 256 & 1 & - \\
        Heat flux    & RNN          & \num{0.0005832402353763642}& \num{0.0005098360860369298}  & \num{0.35173828840625054}& 256 & 1 & - \\
        Heat flux    & LSTM         & \num{0.002400530915105331} & \num{1.329854476833654e-06}  & \num{0.1588930944466614} &  32 & 1 & - \\
        Heat flux    & BiLSTM       & \num{0.009564426449720859} & \num{1.011916973003208e-06}  & \num{0.10662382748174096}&  64 & 1 & - \\
        Heat flux    & TCN          & \num{0.0006384310717849545}& \num{0.0008971430817679988}  & \num{0.1714155590120245} & 256 & 1 & - \\
        Heat flux    & Transformer  & \num{0.0011867023091466015}& \num{0.00027976109820828656} & \num{0.4377518157942493} &  64 & 1 & 4 \\
        \bottomrule
    \end{tabular}%
    }
    \label{tab:params_run4}
\end{table}

\begin{table}[h!]
    \centering
    \caption{Best hyperparameters for dataset RUN5.}
    \resizebox{\textwidth}{!}{%
    \begin{tabular}{@{}llccccccc@{}}
        \toprule
        Target & Model & Learning Rate & Weight Decay & Dropout  & Hidden Units & Layers & Att. Heads \\
        \midrule
        Temperature  & GRU          & \num{0.0008178051930051754}& \num{1.5231388209566893e-05} & \num{0.20216640711315267}& 256 & 4 & - \\
        Temperature  & RNN          & \num{0.00022518662763234933}& \num{2.1671322855110322e-05} & \num{0.17738977435017128}& 128 & 1 & - \\
        Temperature  & LSTM         & \num{0.00607460147116595}  & \num{0.00015010332374947483} & \num{0.39792270977491695}&  32 & 1 & - \\
        Temperature  & BiLSTM       & \num{0.0012668417477211353}& \num{3.0459661849825335e-06} & \num{0.1347433310734189} & 256 & 1 & - \\
        Temperature  & TCN          & \num{0.00048163652412918983}& \num{0.0009320380853660729}  & \num{0.37546452478710046}& 128 & 1 & - \\
        Temperature  & Transformer  & \num{0.00038581991975515976}& \num{0.0002461447025111987}  & \num{0.14812349092566657}& 256 & 1 & 2 \\
        Heat flux    & GRU          & \num{0.004736455535610827} & \num{2.2839293310973253e-05} & \num{0.4016621286598453} & 128 & 1 & - \\
        Heat flux    & RNN          & \num{0.0011717783883076237}& \num{5.311171736498438e-05}  & \num{0.19428415879034064}& 128 & 1 & - \\
        Heat flux    & LSTM         & \num{0.00955102447527712}  & \num{7.878497978472102e-06}  & \num{0.45514592673433196}& 128 & 1 & - \\
        Heat flux    & BiLSTM       & \num{0.0035872627959795023}& \num{0.0002683752206704228}  & \num{0.3780293331468274} & 128 & 1 & - \\
        Heat flux    & TCN          & \num{0.007194314474450196} & \num{3.6727439328111355e-05} & \num{0.10318279717272084}&  32 & 9 & - \\
        Heat flux    & Transformer  & \num{0.0008661354645213036}& \num{2.3258243636816468e-05} & \num{0.33719863301948577}&  64 & 1 & 4 \\
        \bottomrule
    \end{tabular}%
    }
    \label{tab:params_run5}
\end{table}

\begin{table}[h!]
    \centering
    \caption{Best hyperparameters for dataset RUN6.}
    \resizebox{\textwidth}{!}{%
    \begin{tabular}{@{}llccccccc@{}}
        \toprule
        Target & Model & Learning Rate & Weight Decay & Dropout  & Hidden Units & Layers & Att. Heads \\
        \midrule
        Temperature  & GRU          & \num{0.00012485128234516983}& \num{1.1437896460292687e-06} & \num{0.13306208077676113}& 256 & 1 & - \\
        Temperature  & RNN          & \num{0.005521891471225526} & \num{0.0004910626287419497}  & \num{0.4066660666264261} &  64 & 1 & - \\
        Temperature  & LSTM         & \num{0.0002858588012604262}& \num{7.116698955266078e-06}  & \num{0.4052102388676533} & 128 & 1 & - \\
        Temperature  & BiLSTM       & \num{0.008287327601599982} & \num{4.696875794634744e-06}  & \num{0.3139725501247803} & 256 & 1 & - \\
        Temperature  & TCN          & \num{0.0010291791999303503}& \num{0.00034353136003904746} & \num{0.23614345178928312}& 128 & 1 & - \\
        Temperature  & Transformer  & \num{1.3799220740625436e-05}& \num{4.964242443816024e-06}  & \num{0.1456955307143003} & 256 & 7 & 2 \\
        Heat flux    & GRU          & \num{0.0010922801558591168}& \num{0.00021688526477762815} & \num{0.13763044886003928}& 128 & 2 & - \\
        Heat flux    & RNN          & \num{0.0006263914532361203}& \num{1.7980960730872004e-06} & \num{0.3792888971976044} & 256 & 1 & - \\
        Heat flux    & LSTM         & \num{0.0011293887116266138}& \num{0.0008845266424227336}  & \num{0.1001875507341874} & 128 & 1 & - \\
        Heat flux    & BiLSTM       & \num{0.009017106264033549} & \num{2.1166552713207163e-05} & \num{0.20671918924093433}&  64 & 1 & - \\
        Heat flux    & TCN          & \num{0.004382780089881771} & \num{0.000372182028064267}   & \num{0.2259044759513396} & 128 & 5 & - \\
        Heat flux    & Transformer  & \num{0.0006940197229356117}& \num{1.017058988211667e-06}  & \num{0.18356236265725712}& 256 & 1 & 4 \\
        \bottomrule
    \end{tabular}%
    }
    \label{tab:params_run6}
\end{table}

\begin{table}[h!]
    \centering
    \caption{Best hyperparameters for dataset RUN7.}
    \resizebox{\textwidth}{!}{%
    \begin{tabular}{@{}llccccccc@{}}
        \toprule
        Target & Model & Learning Rate & Weight Decay & Dropout  & Hidden Units & Layers & Att. Heads \\
        \midrule
        Temperature  & GRU          & \num{0.00017332679465409439} & \num{0.00036184319549901074} & \num{0.39787133402441005} & 256 & 3  & - \\
        Temperature  & RNN          & \num{0.0017179574934594046} & \num{8.45409011277233e-06}   & \num{0.498939203550265}   &  64 & 1  & - \\
        Temperature  & LSTM         & \num{0.009386383901837412}  & \num{2.9923962599014192e-05} & \num{0.10599441951983885} &  32 & 1  & - \\
        Temperature  & BiLSTM       & \num{0.0010074519011185601} & \num{0.0007230845328561457}  & \num{0.20084007870084963} & 256 & 3  & - \\
        Temperature  & TCN          & \num{0.0005366327314755841} & \num{1.9847879500305437e-06} & \num{0.45490578883560273} & 128 & 1  & - \\
        Temperature  & Transformer  & \num{2.218977258642204e-05} & \num{4.140443268701124e-05}  & \num{0.1290612893592319}  & 128 & 10 & 2 \\
        Heat flux    & GRU          & \num{0.009560117145644894} & \num{2.038100955986585e-06}  & \num{0.3862799381858239} &  32 & 1  & - \\
        Heat flux    & RNN          & \num{0.006902475397868325}  & \num{0.0009760826021568978}  & \num{0.35840076932141435} & 128 & 1  & - \\
        Heat flux    & LSTM         & \num{0.0005421161554088003} & \num{1.3672942195988253e-06} & \num{0.49648982707129563} & 128 & 1  & - \\
        Heat flux    & BiLSTM       & \num{0.008405234686442995}  & \num{0.00015350306265998423} & \num{0.36785270675943776} & 128 & 1  & - \\
        Heat flux    & TCN          & \num{0.0001915500771349203} & \num{3.6395829477919008e-06} & \num{0.17811469447854217} & 256 & 1  & - \\
        Heat flux    & Transformer  & \num{0.00033959732445252926}& \num{0.0009563727755198641}  & \num{0.3554151912530703}  & 256 & 1  & 4 \\
        \bottomrule
    \end{tabular}%
    }
    \label{tab:params_run7}
\end{table}

\begin{table}[h!]
    \centering
    \caption{Best hyperparameters for dataset RUN8.}
    \resizebox{\textwidth}{!}{%
    \begin{tabular}{@{}llccccccc@{}}
        \toprule
        Target & Model & Learning Rate & Weight Decay & Dropout  & Hidden Units & Layers & Att. Heads \\
        \midrule
        Temperature  & GRU          & \num{4.972604789557873e-05}& \num{2.184108542468967e-06}  & \num{0.32317869056694337} & 128 & 4 & - \\
        Temperature  & RNN          & \num{0.002644316209353332} & \num{0.0006978556003526016}  & \num{0.4113858021684518}  & 256 & 1 & - \\
        Temperature  & LSTM         & \num{0.002237368284480792} & \num{9.300393945003841e-05}  & \num{0.34750856436667743} & 128 & 1 & - \\
        Temperature  & BiLSTM       & \num{0.0059990947268350045}& \num{0.000440913585685763}   & \num{0.38917368935075053} & 128 & 1 & - \\
        Temperature  & TCN          & \num{0.002411299318748334} & \num{1.3004116472766339e-05} & \num{0.43115811669083093} & 128 & 1 & - \\
        Temperature  & Transformer  & \num{2.30997200623185e-05} & \num{0.00020359516582249033} & \num{0.12052732206431643} & 128 & 1 & 2 \\
        Heat flux    & GRU          & \num{0.009842735759298943} & \num{0.000983150979171227}   & \num{0.14188442096062376} &  32 & 1 & - \\
        Heat flux    & RNN          & \num{0.00095554627929348}  & \num{6.8765248205395e-05}    & \num{0.3279744738488884}  & 128 & 1 & - \\
        Heat flux    & LSTM         & \num{0.0012583515045391307}& \num{0.00020769482074444012} & \num{0.35335595899706357} &  64 & 1 & - \\
        Heat flux    & BiLSTM       & \num{0.009719020332618634} & \num{1.4428101961560515e-05} & \num{0.10028862979153076} & 256 & 1 & - \\
        Heat flux    & TCN          & \num{0.00488603084778453}  & \num{2.184855046206186e-06}  & \num{0.15835737972697078} &  32 & 1 & - \\
        Heat flux    & Transformer  & \num{0.0008804864327026966}& \num{5.040699085461037e-05}  & \num{0.20340131768377875} &  64 & 1 & 4 \\
        \bottomrule
    \end{tabular}%
    }
    \label{tab:params_run8}
\end{table}

\begin{table}[h!]
    \centering
    \caption{Best hyperparameters for dataset RUN9.}
    \resizebox{\textwidth}{!}{%
    \begin{tabular}{@{}llccccccc@{}}
        \toprule
        Target & Model & Learning Rate & Weight Decay & Dropout  & Hidden Units & Layers & Att. Heads \\
        \midrule
        Temperature  & GRU          & \num{0.0011709430886683485}& \num{6.703698763333816e-05}  & \num{0.1106027571608543}  & 256 & 1 & - \\
        Temperature  & RNN          & \num{0.004817025309092867} & \num{9.057199237184786e-06}  & \num{0.2401930155030527}  &  64 & 1 & - \\
        Temperature  & LSTM         & \num{0.0007108393270358778}& \num{2.5895413188559904e-06} & \num{0.39307011622316274} & 256 & 1 & - \\
        Temperature  & BiLSTM       & \num{0.000986705471949019} & \num{1.2815527848542566e-06} & \num{0.33362076571289206} & 256 & 1 & - \\
        Temperature  & TCN          & \num{0.0003985417887653289}& \num{0.0001801793199060978}  & \num{0.44604477209214766} & 256 & 1 & - \\
        Temperature  & Transformer  & \num{1.4933949499988763e-05}& \num{8.811839142760799e-05}  & \num{0.10738216465767804} &  32 & 1  & 4 \\
        Heat flux    & GRU          & \num{0.0027992404275763026}& \num{4.029560049708701e-05}  & \num{0.14608216708134902} &  64 & 2 & - \\
        Heat flux    & RNN          & \num{0.005627411208900436} & \num{0.0007777990195154621}  & \num{0.36814970209969455} & 128 & 1 & - \\
        Heat flux    & LSTM         & \num{0.006771703074811305} & \num{1.0433190508187189e-06} & \num{0.19060999819731883} &  64 & 1 & - \\
        Heat flux    & BiLSTM       & \num{0.0030507080733531405}& \num{1.3477668803844168e-06} & \num{0.3807788768143813}  & 128 & 1 & - \\
        Heat flux    & TCN          & \num{0.004949135050963912} & \num{0.00040401841529601857} & \num{0.3973280161243682}  &  64 & 10 & - \\
        Heat flux    & Transformer  & \num{4.620278477066097e-05}& \num{1.081536634526999e-06}  & \num{0.10194153927072006} &  64 & 11 & 4 \\
        \bottomrule
    \end{tabular}%
    }
    \label{tab:params_run9}
\end{table}

\begin{table}[h!]
    \centering
    \caption{Best hyperparameters for dataset RUN10.}
    \resizebox{\textwidth}{!}{%
    \begin{tabular}{@{}llccccccc@{}}
        \toprule
        Target & Model & Learning Rate & Weight Decay & Dropout  & Hidden Units & Layers & Att. Heads \\
        \midrule
        Temperature  & GRU          & \num{0.009904853388029766} & \num{0.00014581990558662947} & \num{0.24912362146794687}&  64 & 1 & - \\
        Temperature  & RNN          & \num{0.0017439365641442135}& \num{3.2832748038504627e-05} & \num{0.22186606075318444}& 256 & 1 & - \\
        Temperature  & LSTM         & \num{0.0017001998299054633}& \num{3.274088920634154e-06}  & \num{0.17993492901839647}&  64 & 1 & - \\
        Temperature  & BiLSTM       & \num{0.001688988013166477} & \num{0.0004250550335392657}  & \num{0.49064242718158074}& 128 & 1 & - \\
        Temperature  & TCN          & \num{0.003766503839691325} & \num{2.257213502685221e-05}  & \num{0.4319031565784468} &  32 & 5 & - \\
        Temperature  & Transformer  & \num{3.383151618442306e-05} & \num{0.0005235581179246607}  & \num{0.13753125007298217} & 128 & 1 & 2 \\
        Heat flux    & GRU          & \num{0.0031155480663097625}& \num{5.329394414081454e-06}  & \num{0.1502669179073086} & 256 & 2 & - \\
        Heat flux    & RNN          & \num{0.002273183968367641} & \num{1.5127224753144563e-05} & \num{0.38165196169333543}& 128 & 1 & - \\
        Heat flux    & LSTM         & \num{0.004321305721422249} & \num{1.3214205661079528e-05} & \num{0.18992007352294588}&  64 & 1 & - \\
        Heat flux    & BiLSTM       & \num{0.0019329827193865993}& \num{0.00010015944651841636} & \num{0.2526900490947882} & 128 & 1 & - \\
        Heat flux    & TCN          & \num{0.002989120362377399} & \num{1.1568392384863688e-06} & \num{0.4890248353084064} & 256 & 1 & - \\
        Heat flux    & Transformer  & \num{0.001438931868144639} & \num{6.725596190868635e-05}  & \num{0.2341513494540864} &  32 & 1 & 4 \\
        \bottomrule
    \end{tabular}%
    }
    \label{tab:params_run10}
\end{table}

\begin{table}[h!]
    \centering
    \caption{Best hyperparameters for dataset RUN11.}
    \resizebox{\textwidth}{!}{%
    \begin{tabular}{@{}llccccccc@{}}
        \toprule
        Target & Model & Learning Rate & Weight Decay & Dropout  & Hidden Units & Layers & Att. Heads \\
        \midrule
        Temperature  & GRU          & \num{0.000827666619536676} & \num{3.65124119246989e-06}   & \num{0.1799294788823506} & 256 & 1 & - \\
        Temperature  & RNN          & \num{0.00220091267189474}  & \num{2.1953319870384738e-05} & \num{0.14908332088014734}& 256 & 1 & - \\
        Temperature  & LSTM         & \num{0.0027784558009063404}& \num{2.3842385990823688e-06} & \num{0.12936424760465523}&  64 & 1 & - \\
        Temperature  & BiLSTM       & \num{0.009932984555595742} & \num{0.00016477100644191788} & \num{0.1462726177951475} & 256 & 1 & - \\
        Temperature  & TCN          & \num{0.004399476705885281} & \num{2.2336581462790977e-06} & \num{0.47301088179947176} & 256 & 1 & - \\
        Temperature  & Transformer  & \num{0.0004062585609351196} & \num{0.0004462990207649082}  & \num{0.10433154361201386} &  64 & 3 & 2 \\
        Heat flux    & GRU          & \num{0.002666525721491688} & \num{2.682114896815709e-05}  & \num{0.1967469913175541} & 128 & 1 & - \\
        Heat flux    & RNN          & \num{0.00022773664943179255}& \num{2.0741630118275168e-05} & \num{0.41633072212426414}& 256 & 3 & - \\
        Heat flux    & LSTM         & \num{0.005658089948675763} & \num{9.111599393168711e-05}  & \num{0.31575209328492837}&  32 & 1 & - \\
        Heat flux    & BiLSTM       & \num{0.004084593378760253} & \num{4.726862543910042e-05}  & \num{0.20731214260178968}& 128 & 1 & - \\
        Heat flux    & TCN          & \num{0.0016746128631097495}& \num{0.00011179084263350548} & \num{0.3908314575814597} & 128 & 1 & - \\
        Heat flux    & Transformer  & \num{0.00034241074677981504}& \num{4.7209779258356485e-06} & \num{0.23017450499787542}& 128 & 1 & 4 \\
        \bottomrule
    \end{tabular}%
    }
    \label{tab:params_run11}
\end{table}

\begin{table}[h!]
    \centering
    \caption{Best hyperparameters for dataset RUN12.}
    \resizebox{\textwidth}{!}{%
    \begin{tabular}{@{}llccccccc@{}}
        \toprule
        Target & Model & Learning Rate & Weight Decay & Dropout  & Hidden Units & Layers & Att. Heads \\
        \midrule
        Temperature  & GRU          & \num{0.0006467191634695095}& \num{5.415300792163225e-05}  & \num{0.3653322694429621} &  64 & 1 & - \\
        Temperature  & RNN          & \num{0.0007867153370151472}& \num{0.0002792407749024872}  & \num{0.21115032113135862}& 128 & 1 & - \\
        Temperature  & LSTM         & \num{0.003643864061992927} & \num{2.492181876391826e-06}  & \num{0.29791439668942}   &  64 & 1 & - \\
        Temperature  & BiLSTM       & \num{0.0017524500162043103}& \num{3.2859572811876804e-06} & \num{0.36318378248184036}& 128 & 1 & - \\
        Temperature  & TCN          & \num{0.0007258617980290104}& \num{3.4453717410266247e-06} & \num{0.13028769793254663}& 128 & 1 & - \\
        Temperature  & Transformer  & \num{0.00012181443967415132}& \num{1.873726292985057e-05} & \num{0.18633802333514282}& 128 & 1 & 2 \\
        Heat flux    & GRU          & \num{0.003693231866522231} & \num{7.202147672625853e-06}  & \num{0.2440627388742207} &  64 & 1 & - \\
        Heat flux    & RNN          & \num{0.0008380504804340783}& \num{9.873676436244347e-05}  & \num{0.2285948878945837} & 128 & 8 & - \\
        Heat flux    & LSTM         & \num{0.00452898822995025}  & \num{1.28648575156968e-05}   & \num{0.422799877619952}  & 128 & 1 & - \\
        Heat flux    & BiLSTM       & \num{0.004980320717576035} & \num{0.00023230391987134134} & \num{0.1955625865706011} & 128 & 1 & - \\
        Heat flux    & TCN          & \num{0.002705005088699739} & \num{0.0009476214483669777}  & \num{0.10710162658690382}& 128 & 1 & - \\
        Heat flux    & Transformer  & \num{0.0002459423859848168}& \num{4.434083446686095e-05}  & \num{0.22233822384347227}& 128 & 1 & 2 \\
        \bottomrule
    \end{tabular}%
    }
    \label{tab:params_run12}
\end{table}

\FloatBarrier

\subsection{Generalised Temperature Predictors}

\begin{table}[h!]
    \centering
    \caption{Best hyperparameters for held-out dataset RUN1.}
    \resizebox{\textwidth}{!}{%
        \begin{tabular}{@{}llcccccc@{}}
        \toprule
        Target & Model & Learning Rate & Weight Decay & Dropout  & Hidden Units & Layers & Att. Heads \\
        \midrule
        Temperature  & GRU          & \num{0.00040979931054102516} & \num{0.0007408313750656381} & \num{0.4746767861503724} & 256 & 1 & - \\
        Temperature  & RNN          & \num{0.0001063641051026175}  & \num{0.00012813600405783978}& \num{0.12838109512953597}&  64 & 1 & - \\
        Temperature  & LSTM         & \num{0.007065114679296662}   & \num{8.416936913049958e-05} & \num{0.33730077752033455}& 256 & 1 & - \\
        Temperature  & TCN          & \num{0.00017249935253865506} & \num{2.1488617145678955e-06}& \num{0.11007861469916191}& 256 & 1 & - \\
        Temperature  & BiLSTM       & \num{0.004503869693059496}   & \num{6.217323569347632e-05} & \num{0.28060214167606434}& 256 & 1 & - \\
        Temperature  & Transformer  & \num{0.0001764662353045022}  & \num{2.9032109318433946e-05}& \num{0.31033502672268953}& 256 & 3 & 2 \\
        Heat flux    & GRU          & \num{0.0020294154869301764}  & \num{1.040611325432112e-06} & \num{0.18367763207437138}&  64 & 1 & - \\
        Heat flux    & RNN          & \num{0.001184152117458863}   & \num{1.1842729539247798e-06}& \num{0.180663409132315}  & 128 & 1 & - \\
        Heat flux    & LSTM         & \num{0.00021520483414101717} & \num{6.0067690278845415e-06}& \num{0.1974441242551644} & 256 & 1 & - \\
        Heat flux    & TCN          & \num{0.0011562923813997965}  & \num{1.0815620852246626e-05}& \num{0.40680472054610906}&  64 & 1 & - \\
        Heat flux    & BiLSTM       & \num{0.0019541072551392162}  & \num{3.965577956492565e-06} & \num{0.14284728405260366}&  64 & 1 & - \\
        Heat flux    & Transformer  & \num{2.8407569082710495e-05} & \num{4.512093156453251e-05} & \num{0.22578040098710078}& 256 & 1 & 2 \\
        \bottomrule
        \end{tabular}%
    }
    \label{tab:gen_params_run1}
\end{table}

\begin{table}[h!]
    \centering
    \caption{Best hyperparameters for held-out dataset RUN2.}
    \resizebox{\textwidth}{!}{%
        \begin{tabular}{@{}llccccccc@{}}
        \toprule
        Target & Model & Learning Rate & Weight Decay & Dropout  & Hidden Units & Layers & Att. Heads \\
        \midrule
        Temperature  & GRU          & \num{0.00040979931054102516} & \num{0.0007408313750656381} & \num{0.4746767861503724} & 256 & 1 & - \\
        Temperature  & RNN          & \num{0.0001063641051026175}  & \num{0.00012813600405783978}& \num{0.12838109512953597}&  64 & 1 & - \\
        Temperature  & LSTM         & \num{0.007065114679296662}   & \num{8.416936913049958e-05} & \num{0.33730077752033455}& 256 & 1 & - \\
        Temperature  & TCN          & \num{0.00017249935253865506} & \num{2.1488617145678955e-06}& \num{0.11007861469916191}& 256 & 1 & - \\
        Temperature  & BiLSTM       & \num{0.004503869693059496}   & \num{6.217323569347632e-05} & \num{0.28060214167606434}& 256 & 1 & - \\
        Temperature  & Transformer  & \num{0.0001764662353045022}  & \num{2.9032109318433946e-05}& \num{0.31033502672268953}& 256 & 3 & 2 \\
        Heat flux    & GRU          & \num{0.0010833799249610684}  & \num{2.411658142536657e-06} & \num{0.26830241831805096}&  64 & 1 & - \\
        Heat flux    & RNN          & \num{0.00819407640158193}    & \num{4.901521968105056e-06} & \num{0.23911788974987944}&  32 & 1 & - \\
        Heat flux    & LSTM         & \num{0.0003031557726330065}  & \num{5.6043806365056725e-06}& \num{0.32587819411059055}&  32 & 1 & - \\
        Heat flux    & TCN          & \num{0.0014082402151201477}  & \num{4.550504616078974e-06} & \num{0.3802291074302211} &  32 & 3 & - \\
        Heat flux    & BiLSTM       & \num{0.0007647273309899475}  & \num{2.1967082708369638e-05} & \num{0.4874035208727249} &  32 & 1 & - \\
        Heat flux    & Transformer  & \num{0.002571552288517391}   & \num{0.0003385600796790919} & \num{0.40605052012698073}&  64 & 2 & 2 \\
        \bottomrule
        \end{tabular}%
    }
    \label{tab:gen_params_run2}
\end{table}

\begin{table}[h!]
    \centering
    \caption{Best hyperparameters for held-out dataset RUN3.}
    \resizebox{\textwidth}{!}{%
        \begin{tabular}{@{}llccccccc@{}}
        \toprule
        Target & Model & Learning Rate & Weight Decay & Dropout  & Hidden Units & Layers & Att. Heads \\
        \midrule
        Temperature  & GRU          & \num{0.000258342061821624}  & \num{7.068595547047084e-05}  & \num{0.342420898264997}  &  64 & 1  & - \\
        Temperature  & RNN          & \num{0.0007963306610981715} & \num{7.99430100003459e-05}   & \num{0.40328475591706003}& 128 & 1  & - \\
        Temperature  & LSTM         & \num{0.001039557143447382}  & \num{0.0006093764677985728}  & \num{0.26092644416071864}&  32 & 1  & - \\
        Temperature  & TCN          & \num{0.0014412403844862238} & \num{1.6806797654938575e-05} & \num{0.32316774425045514}& 128 & 10 & - \\
        Temperature  & BiLSTM       & \num{0.00040505774368409424}& \num{1.1851202580765715e-06} & \num{0.2180300884570598} &  32 & 1  & - \\
        Temperature  & Transformer  & \num{0.0007621782567727604} & \num{0.0009025425851579899}  & \num{0.4593395911925804} &  32 & 1  & 4 \\
        Heat flux    & GRU          & \num{0.00011147973690952839}& \num{7.620374825192245e-05}  & \num{0.49674769857092477}& 128 & 1  & - \\
        Heat flux    & RNN          & \num{0.0002785180718542163} & \num{1.1471073720320402e-05} & \num{0.48138298757251785}& 128 & 1  & - \\
        Heat flux    & LSTM         & \num{0.0001224274078593802} & \num{1.188299843366632e-06}  & \num{0.3886495260880816} & 256 & 1  & - \\
        Heat flux    & TCN          & \num{0.0005029716354718202} & \num{2.3097879385998905e-05} & \num{0.27949349936522794}& 256 & 12 & - \\
        Heat flux    & BiLSTM       & \num{0.000939891765946917}  & \num{0.000918191654871912}   & \num{0.27006728079362285}& 128 & 1  & - \\
        Heat flux    & Transformer  & \num{0.00096878935823489}   & \num{4.1470745360884464e-05} & \num{0.3474564456264885} &  32 & 3  & 4 \\
        \bottomrule
        \end{tabular}%
    }
    \label{tab:gen_params_run3}
\end{table}

\begin{table}[h!]
    \centering
    \caption{Best hyperparameters for held-out dataset RUN4}
    \resizebox{\textwidth}{!}{%
        \begin{tabular}{@{}llccccccc@{}}
        \toprule
        Target & Model & Learning Rate & Weight Decay & Dropout & Hidden Units & Layers & Att. Heads \\
        \midrule
        Temperature  & GRU          & \num{0.0015814754333131015} & \num{3.406861526453605e-06}  & \num{0.3111155417920706} & 256 & 1 & - \\
        Temperature  & RNN          & \num{0.00012245796976350786}& \num{1.157798879781557e-06}  & \num{0.3021263060945075} & 256 & 1 & - \\
        Temperature  & LSTM         & \num{0.009668676584534236}  & \num{1.1309996391769985e-06} & \num{0.4990973279595636} & 128 & 1 & - \\
        Temperature  & TCN          & \num{0.001120975811579074}  & \num{1.5682322688319198e-06} & \num{0.48579913178831424}& 128 & 8 & - \\
        Temperature  & BiLSTM       & \num{0.0006559340739917502} & \num{2.0167660461184114e-05} & \num{0.12204261499931268}& 128 & 1 & - \\
        Temperature  & Transformer  & \num{0.0005295282671940952} & \num{0.00017754039766480425} & \num{0.18448244850501835}&  32 & 3 & 4 \\
        Heat flux    & GRU          & \num{0.006633965527947345}  & \num{0.00012192329902893836} & \num{0.13035870403469266}&  32 & 1 & - \\
        Heat flux    & RNN          & \num{6.983924752848209e-05} & \num{2.8055383523369165e-06} & \num{0.4560801864329772} & 256 & 2 & - \\
        Heat flux    & LSTM         & \num{0.00027625251448962457}& \num{0.00038485030027016426} & \num{0.31883149540000894}&  64 & 1 & - \\
        Heat flux    & TCN          & \num{0.00010814762656736621}& \num{4.165468079233974e-06}  & \num{0.3719132833037321} &  64 & 8 & - \\
        Heat flux    & BiLSTM       & \num{0.00037647713136810623}& \num{0.0008057271285900306}  & \num{0.4516368740786504} &  64 & 1 & - \\
        Heat flux    & Transformer  & \num{0.000719250870513009}  & \num{1.0761509334794149e-06} & \num{0.10180386617291914}& 256 & 1 & 4 \\
        \bottomrule
        \end{tabular}%
    }
    \label{tab:gen_params_run4}
\end{table}

\begin{table}[h!]
    \centering
    \caption{Best hyperparameters for held-out dataset RUN5}
    \resizebox{\textwidth}{!}{%
        \begin{tabular}{@{}llccccccc@{}}
        \toprule
        Target & Model & Learning Rate & Weight Decay & Dropout & Hidden Units & Layers & Att. Heads \\
        \midrule
        Temperature  & GRU          & \num{0.0009806890223803628} & \num{0.0001553903782495859} & \num{0.10545124964630186} &  64 & 1 & - \\
        Temperature  & RNN          & \num{5.9248019396948084e-05}& \num{5.689767922978785e-05} & \num{0.19121076018825167} & 128 & 1 & - \\
        Temperature  & LSTM         & \num{0.0036808847970880953} & \num{0.00029255774150022785}& \num{0.18485227128165868} & 128 & 1 & - \\
        Temperature  & TCN          & \num{0.0008411405854993846} & \num{6.676593423848363e-05} & \num{0.35196601938691596} &  64 & 2 & - \\
        Temperature  & BiLSTM       & \num{0.0003676749532837029} & \num{1.7582958622568826e-05}& \num{0.395768138918959}   &  64 & 1 & - \\
        Temperature  & Transformer  & \num{0.0005004648579118232} & \num{7.1799349341596585e-06}& \num{0.2854871692837096}  &  32 & 3 & 2 \\
        Heat flux    & GRU          & \num{0.000338755177110714}  & \num{6.608635197279926e-06} & \num{0.4217295225278407} & 256 & 1 & - \\
        Heat flux    & RNN          & \num{0.000347053110434319}  & \num{6.1623672391649e-05}   & \num{0.30178151583871804}&  64 & 1 & - \\
        Heat flux    & LSTM         & \num{0.0008297730490990398} & \num{2.135976064463516e-06} & \num{0.3795212825962797} &  64 & 1 & - \\
        Heat flux    & TCN          & \num{0.0010673845263674243} & \num{8.07675728249059e-05}  & \num{0.16953941091531755}&  32 & 11 & - \\
        Heat flux    & BiLSTM       & \num{0.0005506642830934857} & \num{0.0002511263155130295} & \num{0.3615863625989668} & 128 & 1 & - \\
        Heat flux    & Transformer  & \num{0.00010090238451581185}& \num{7.195984481132496e-05} & \num{0.20240068995626498}&  64 & 1 & 4 \\
        \bottomrule
        \end{tabular}%
    }
    \label{tab:gen_params_run5}
\end{table}

\begin{table}[h!]
    \centering
    \caption{Best hyperparameters for held-out dataset RUN6}
    \resizebox{\textwidth}{!}{%
        \begin{tabular}{@{}llccccccc@{}}
        \toprule
        Target & Model & Learning Rate & Weight Decay & Dropout & Hidden Units & Layers & Att. Heads \\
        \midrule
        Temperature  & GRU          & \num{0.00931541471526421}   & \num{0.0001430011711231818} & \num{0.13741748607850285} & 128 & 1 & - \\
        Temperature  & RNN          & \num{0.00026562586735298925}& \num{6.233162661823234e-05} & \num{0.1894661944790259}  & 256 & 1 & - \\
        Temperature  & LSTM         & \num{0.0008346713070694929} & \num{7.857184910837286e-06} & \num{0.40885620553260493} & 256 & 1 & - \\
        Temperature  & TCN          & \num{0.0012185443678860717} & \num{8.591620514224261e-05} & \num{0.112626619794657}   &  64 & 11 & - \\
        Temperature  & BiLSTM       & \num{0.0009665366416177326} & \num{5.117301833393697e-06} & \num{0.46782550068342077} &  64 & 1 & - \\
        Temperature  & Transformer  & \num{9.431955075565503e-05} & \num{0.00011507635799295004}& \num{0.1052640000099564}  &  32 & 1 & 4 \\
        Heat flux    & GRU          & \num{0.00013376767921223867}& \num{6.000075594011579e-06} & \num{0.3303863537153569}  & 128 & 1 & - \\
        Heat flux    & RNN          & \num{0.000771388279340839}  & \num{1.0803553902035182e-05}& \num{0.49993723492981074} &  32 & 1 & - \\
        Heat flux    & LSTM         & \num{0.0003254353694858945} & \num{7.3494802193227924e-06}& \num{0.13382557266726916} & 256 & 1 & - \\
        Heat flux    & TCN          & \num{0.00096864126744974}   & \num{1.238048329351028e-05} & \num{0.2113252983389822}  &  64 & 12 & - \\
        Heat flux    & BiLSTM       & \num{0.005873711969270803}  & \num{1.2408384899209727e-05}& \num{0.4163614915025014}  &  64 & 1 & - \\
        Heat flux    & Transformer  & \num{0.0002680974992201504} & \num{0.0003084004262959043} & \num{0.18013404735139393} &  64 & 2 & 4 \\
        \bottomrule
        \end{tabular}%
    }
    \label{tab:gen_params_run6}
\end{table}

\begin{table}[h!]
    \centering
    \caption{Best hyperparameters for held-out dataset RUN7}
    \resizebox{\textwidth}{!}{%
        \begin{tabular}{@{}llccccccc@{}}
        \toprule
        Target & Model & Learning Rate & Weight Decay & Dropout & Hidden Units & Layers & Att. Heads \\
        \midrule
        Temperature  & GRU          & \num{0.0020778511127647952} & \num{1.0712121973436172e-06} & \num{0.26379600348187804}& 128 & 1 & - \\
        Temperature  & RNN          & \num{0.0002725598264857259} & \num{0.0007992829699454149}  & \num{0.4857685328685313} & 128 & 1 & - \\
        Temperature  & LSTM         & \num{0.00018182732508240473}& \num{1.1756661139236157e-05} & \num{0.10323366909040421}& 128 & 1 & - \\
        Temperature  & TCN          & \num{0.00035605336797066403}& \num{0.00034739652297836735} & \num{0.19102084823013268}& 128 & 3 & - \\
        Temperature  & BiLSTM       & \num{0.000152544872968003}  & \num{9.124019922861815e-05}  & \num{0.3317652887762338} & 128 & 1 & - \\
        Temperature  & Transformer  & \num{0.0001295377351116925} & \num{0.00018778304316511185} & \num{0.48949146703293633}& 256 & 2 & 4 \\
        Heat flux    & GRU          & \num{0.0006798027111421682} & \num{1.0088462112611216e-06} & \num{0.2807272196121186} &  32 & 1 & - \\
        Heat flux    & RNN          & \num{0.00018636884805158904}& \num{7.05800345775207e-06}   & \num{0.42601227666055813}&  32 & 1 & - \\
        Heat flux    & LSTM         & \num{0.0017208802609697813} & \num{2.267246341710859e-06}  & \num{0.2257848448975535} &  32 & 1 & - \\
        Heat flux    & TCN          & \num{0.00042433750810532593}& \num{3.4782670542910138e-06} & \num{0.21742858697438777}&  32 & 5 & - \\
        Heat flux    & BiLSTM       & \num{0.000610628848157197}  & \num{0.00013138142913937453} & \num{0.3797665515287416} &  32 & 1 & - \\
        Heat flux    & Transformer  & \num{0.000509331123937777}  & \num{1.0164340568941472e-06} & \num{0.42006927969470576}&  32 & 2 & 4 \\
        \bottomrule
        \end{tabular}%
    }
    \label{tab:gen_params_run7}
\end{table}

\begin{table}[h!]
    \centering
    \caption{Best hyperparameters for held-out dataset RUN8}
    \resizebox{\textwidth}{!}{%
        \begin{tabular}{@{}llccccccc@{}}
        \toprule
        Target & Model & Learning Rate & Weight Decay & Dropout & Hidden Units & Layers & Att. Heads \\
        \midrule
        Temperature  & GRU          & \num{0.0016037509970165964} & \num{0.0001499376058758628} & \num{0.49191511866449855}&  64 & 1 & - \\
        Temperature  & RNN          & \num{0.00022693932027764192}& \num{9.603180589277933e-06} & \num{0.3136560850575316} &  64 & 1 & - \\
        Temperature  & LSTM         & \num{0.003000033950663127}  & \num{1.8504959408034007e-05}& \num{0.4213694691429296} &  64 & 1 & - \\
        Temperature  & TCN          & \num{0.00036507934435379524}& \num{1.0590559569905342e-05}& \num{0.18494086501532775}& 128 & 2 & - \\
        Temperature  & BiLSTM       & \num{0.0013393527038446547} & \num{6.421208647246915e-05} & \num{0.479851976564623}  & 256 & 1 & - \\
        Temperature  & Transformer  & \num{0.00019578438768260619}& \num{1.0702206852793355e-06}& \num{0.43924379082682297}&  64 & 1 & 4 \\
        Heat flux    & GRU          & \num{0.0013683826942903102} & \num{1.5867703374185626e-05}& \num{0.29296221794625704}& 128 & 1 & - \\
        Heat flux    & RNN          & \num{0.0006349798224032557} & \num{0.0008412036561192154} & \num{0.3243825381428789} & 128 & 1 & - \\
        Heat flux    & LSTM         & \num{0.0030005377042798295} & \num{1.6220187394257677e-06}& \num{0.1130496352971254} &  64 & 1 & - \\
        Heat flux    & TCN          & \num{0.0002920913532226626} & \num{3.77710187157748e-06}  & \num{0.34425903119857054}& 256 & 4 & - \\
        Heat flux    & BiLSTM       & \num{0.003942714112985374}  & \num{1.8649909139098583e-06}& \num{0.3421656290183821} & 256 & 1 & - \\
        Heat flux    & Transformer  & \num{6.714100210409285e-05}  & \num{1.3909571716861935e-06}& \num{0.39149660163352407}& 128 & 2 & 2 \\
        \bottomrule
        \end{tabular}%
    }
    \label{tab:gen_params_run8}
\end{table}

\begin{table}[h!]
    \centering
    \caption{Best hyperparameters for held-out dataset RUN9}
    \resizebox{\textwidth}{!}{%
        \begin{tabular}{@{}llccccccc@{}}
        \toprule
        Target & Model & Learning Rate & Weight Decay & Dropout & Hidden Units & Layers & Att. Heads \\
        \midrule
        Temperature  & GRU          & \num{0.0006141988616367451} & \num{0.0008556268145211572} & \num{0.25324394518780396}&  32 & 1 & - \\
        Temperature  & RNN          & \num{0.0005826755539706163} & \num{6.1026532435874215e-06}& \num{0.10229431314445006}& 128 & 1 & - \\
        Temperature  & LSTM         & \num{0.006297535785768968}  & \num{4.804338436221907e-06} & \num{0.11376361617939307}& 256 & 1 & - \\
        Temperature  & TCN          & \num{0.001794484951492231}  & \num{4.110029517722537e-06} & \num{0.1105081472680513} &  64 & 9 & - \\
        Temperature  & BiLSTM       & \num{0.00025399195092747767}& \num{0.00018316070985527024}& \num{0.2050339048912294} & 128 & 1 & - \\
        Temperature  & Transformer  & \num{0.0009000719384487974} & \num{1.634070870398791e-05} & \num{0.3765447172716492} & 128 & 3 & 4 \\
        Heat flux    & GRU          & \num{0.0002089914321138388} & \num{4.448578546076982e-05} & \num{0.15297481158251408}& 128 & 1 & - \\
        Heat flux    & RNN          & \num{0.00015454211194262707}& \num{1.503842844873223e-05} & \num{0.31422083547132196}& 256 & 1 & - \\
        Heat flux    & LSTM         & \num{2.2702622953066397e-05}& \num{0.0001376319862521368} & \num{0.23996367647970657}& 256 & 1 & - \\
        Heat flux    & TCN          & \num{0.0008263392220256184} & \num{0.0002666898868727386} & \num{0.11492990143410209}& 256 & 9 & - \\
        Heat flux    & BiLSTM       & \num{0.0015497332586097909} & \num{0.00026314537970878113}& \num{0.2843932968162528} & 128 & 1 & - \\
        Heat flux    & Transformer  & \num{0.00024099112701431318}& \num{1.477055556391151e-05} & \num{0.3183256037164383} &  64 & 2 & 4 \\
        \bottomrule
        \end{tabular}%
    }
    \label{tab:gen_params_run9}
\end{table}

\begin{table}[h!]
    \centering
    \caption{Best hyperparameters for held-out dataset RUN10}
    \resizebox{\textwidth}{!}{%
        \begin{tabular}{@{}llccccccc@{}}
        \toprule
        Target & Model & Learning Rate & Weight Decay & Dropout & Hidden Units & Layers & Att. Heads \\
        \midrule
        Temperature  & GRU          & \num{0.0013259386678280134} & \num{0.00025701049120674174}& \num{0.3321277349627159} & 128 & 1 & - \\
        Temperature  & RNN          & \num{6.938765760945092e-05} & \num{1.7893876890189736e-05}& \num{0.10562304766414454}& 256 & 4 & - \\
        Temperature  & LSTM         & \num{0.0056314567416610685} & \num{1.6723857939128624e-05}& \num{0.11074146148271363}&  32 & 1 & - \\
        Temperature  & TCN          & \num{0.0034935284786027187} & \num{2.5525832547494835e-06}& \num{0.3017147900198891} & 128 & 6 & - \\
        Temperature  & BiLSTM       & \num{0.003508621342183773}  & \num{0.0004760489845122589} & \num{0.10403962878433476}& 128 & 1 & - \\
        Temperature  & Transformer  & \num{0.00033018201304221416}& \num{9.66499064481518e-05}  & \num{0.43712741106906006}& 128 & 2 & 2 \\
        Heat flux    & GRU          & \num{7.796557270913755e-05} & \num{0.0008661005207858792} & \num{0.3796232069277649} & 256 & 1 & - \\
        Heat flux    & RNN          & \num{0.0004878976620451224} & \num{1.1493502092970742e-06}& \num{0.24672161779316304}& 128 & 1 & - \\
        Heat flux    & LSTM         & \num{0.0001655612902035545} & \num{0.0003696815989558295} & \num{0.390817437395774} & 256 & 1 & - \\
        Heat flux    & TCN          & \num{0.0011982219092605323} & \num{6.695299369081865e-05} & \num{0.3138513448740876} &  32 & 4 & - \\
        Heat flux    & BiLSTM       & \num{0.0010100027004540716} & \num{7.195444208108711e-06} & \num{0.13722245360639973}& 256 & 1 & - \\
        Heat flux    & Transformer  & \num{0.0005926552402760135} & \num{7.335145191172009e-05} & \num{0.2400638069708953} & 128 & 1 & 4 \\
        \bottomrule
        \end{tabular}%
    }
    \label{tab:gen_params_run10}
\end{table}

\begin{table}[h!]
    \centering
    \caption{Best hyperparameters for held-out dataset RUN11}
    \resizebox{\textwidth}{!}{%
        \begin{tabular}{@{}llccccccc@{}}
        \toprule
        Target & Model & Learning Rate & Weight Decay & Dropout & Hidden Units & Layers & Att. Heads \\
        \midrule
        Temperature  & GRU          & \num{0.00028400795808472545}& \num{1.2455675722089909e-06} & \num{0.11310723886323687}& 256 & 1 & - \\
        Temperature  & RNN          & \num{0.0026630107499429206} & \num{7.783217208058973e-06}  & \num{0.3658991774243775} &  32 & 1 & - \\
        Temperature  & LSTM         & \num{0.001125797329385664}  & \num{5.996728284155148e-05}  & \num{0.27198690663730224}&  64 & 1 & - \\
        Temperature  & TCN          & \num{0.0012363537131242683} & \num{3.162917662727621e-05}  & \num{0.3403258404588308} &  32 & 9 & - \\
        Temperature  & BiLSTM       & \num{0.00027812784285284484}& \num{7.997558989844265e-06}  & \num{0.3594492753681212} &  32 & 1 & - \\
        Temperature  & Transformer  & \num{0.0001974933272967234} & \num{3.30728654713219e-05}   & \num{0.3408261811320728} &  64 & 1 & 4 \\
        Heat flux    & GRU          & \num{0.0002549359350808764} & \num{0.0001162994124574891} & \num{0.2340103274011306} &  64 & 1 & - \\
        Heat flux    & RNN          & \num{0.0007520348383318458} & \num{0.0004601943035459258} & \num{0.3702105824410882} & 256 & 1 & - \\
        Heat flux    & LSTM         & \num{0.00010528081672830198}& \num{4.69253001744227e-06}   & \num{0.3035171662911475} & 256 & 1 & - \\
        Heat flux    & TCN          & \num{0.0007986630200241093} & \num{0.0007619119645797824} & \num{0.4990504404710702} & 256 & 5 & - \\
        Heat flux    & BiLSTM       & \num{0.002917202740047781}  & \num{1.1685953915468376e-05} & \num{0.39437691149909865}&  32 & 1 & - \\
        Heat flux    & Transformer  & \num{0.00010724890489245594}& \num{3.061278886428066e-05} & \num{0.3362140681623505} &  32 & 1 & 4 \\
        \bottomrule
        \end{tabular}%
    }
    \label{tab:gen_params_run11}
\end{table}

\begin{table}[h!]
    \centering
    \caption{Best hyperparameters for held-out dataset RUN12}
    \resizebox{\textwidth}{!}{%
        \begin{tabular}{@{}llccccccc@{}}
        \toprule
        Target & Model & Learning Rate & Weight Decay & Dropout & Hidden Units & Layers & Att. Heads \\
        \midrule
        Temperature  & GRU          & \num{0.00040913370352637485}& \num{0.00021059409939601085} & \num{0.1981234720990502} & 128 & 1 & - \\
        Temperature  & RNN          & \num{0.00041734654300332863}& \num{5.828585148760363e-06}  & \num{0.38062374441055546}&  64 & 1 & - \\
        Temperature  & LSTM         & \num{0.0010962554408664156} & \num{0.00021319516735487736} & \num{0.3954967741255978} &  64 & 1 & - \\
        Temperature  & TCN          & \num{0.0018118860346453014} & \num{2.83989216886796e-05}   & \num{0.2568705897504502} &  64 & 5 & - \\
        Temperature  & BiLSTM       & \num{0.0015532142957893282} & \num{8.119683793123975e-06}  & \num{0.2815331758674778} & 256 & 1 & - \\
        Temperature  & Transformer  & \num{5.789849540940546e-05} & \num{6.343167678238918e-06}  & \num{0.15753602865232702}& 128 & 2 & 2 \\
        Heat flux    & GRU          & \num{0.004788202763291111}  & \num{8.316784472589584e-06}  & \num{0.24804084871303325}&  32 & 1 & - \\
        Heat flux    & RNN          & \num{0.0007777098430711601} & \num{0.0006993453278884394}  & \num{0.3110524381062886} & 128 & 1 & - \\
        Heat flux    & LSTM         & \num{0.009607272222449613}  & \num{4.19597897443911e-06}   & \num{0.49048531985944194}& 256 & 1 & - \\
        Heat flux    & TCN          & \num{0.0009436761823238677} & \num{8.520263187073402e-06}  & \num{0.14232087493388532}& 256 & 5 & - \\
        Heat flux    & BiLSTM       & \num{0.0038046099286158704} & \num{5.280270071339244e-05}  & \num{0.46148386507556405}& 256 & 1 & - \\
        Heat flux    & Transformer  & \num{0.0001743125130588093} & \num{1.6912599560260312e-05} & \num{0.2620286368415181} & 128 & 3 & 2 \\
        \bottomrule
        \end{tabular}%
    }
    \label{tab:gen_params_run12}
\end{table}

\end{document}